\documentclass[generic,reqno]{imsart}
\usepackage[numbers]{natbib}

\renewcommand\baselinestretch{1}

\RequirePackage[OT1]{fontenc}
\RequirePackage{amsthm,amsmath,natbib}
\RequirePackage[colorlinks,citecolor=blue,urlcolor=blue,breaklinks=true]{hyperref}
\RequirePackage{hypernat}
\RequirePackage{comment, listings}
\RequirePackage{algorithm2e} 
\RequirePackage{algpseudocode,caption}
\RequirePackage{graphicx,subfigure,latexsym,amssymb,mathtools}

\RequirePackage{float,epsfig,multirow,rotating}
\RequirePackage{upgreek}
\RequirePackage[hyphenbreaks]{breakurl}
\startlocaldefs

\numberwithin{equation}{section}
\theoremstyle{plain}

\newtheorem{definition}{Definition}[section]
\newtheorem{remark}{Remark}[section]

\newcommand{\bTheta}{\boldsymbol{\Theta}}

\newcommand{\bSigma}{\boldsymbol{\Sigma}}

\newcommand{\bPsi}{\boldsymbol{\Psi}}

\newcommand{\bV}{\boldsymbol{V}}

\endlocaldefs

\begin{document}
\renewcommand\baselinestretch{1}

\begin{frontmatter}
\title{Individualised recovery trajectories of patients with impeded mobility, using distance between probability distributions of learnt graphs} 


\runtitle{Recovery trajectories using random graphs}

\begin{aug}
\author{                                                                      
{\fnms{Chuqiao} \snm{Zhang}\thanksref{m2}\ead[label=e3]{joe.zhang@brunel.ac.uk}},
  {\fnms{Crina} \snm{Grosan}\thanksref{m1}\ead[label=e1]{crina.grosan@kcl.ac.uk}},
{\fnms{Dalia} \snm{Chakrabarty}\thanksref{m2,m3}\ead[label=e2]{dalia.chakrabarty@york.ac.uk}}
},

\runauthor{Zhang $\&$ Chakrabarty}

\address{\thanksmark{m1} 
Applied Technologies for Clinical Care\\
King's College London,U.K.\\                                                                       
\printead*{e1}                                                                
}
\address{\thanksmark{m2} Department of Mathematics\\
Brunel University London\\
Uxbridge UB8 3PH\\
U.K.\\                                                                        
\printead*{e3}
}
\address{\thanksmark{m3} Department of Mathematics\\
University of York\\
York YO10 5DD\\
U.K.\\                                                                        
\printead*{e2}                                                                
}

\end{aug}

\begin{abstract}
{Patients who are undergoing physical rehabilitation, benefit from feedback
  that follows from reliable assessment of their cumulative performance
  attained at a given time. In
  this paper, we provide a method for the learning of the recovery trajectory
  of an individual patient, as they undertake exercises as part of their
  physical therapy towards recovery of their loss of movement ability,
  following a critical illness. The difference between the Movement Recovery
  Scores (MRSs) attained by a patient, when undertaking a given exercise
  routine on successive instances, is given by a statistical
  distance/divergence between the (posterior) probabilities of random graphs
  that are Bayesianly learnt using time series data on locations of 20 of the
  patient's joints, recorded on an e-platform as the patient exercises. This allows for the computation
  of the MRS on every occasion the patient undertakes this exercise, using
  which, the recovery trajectory is drawn. We learn each graph as a Random
  Geometric Graph drawn in a probabilistic metric space, and identify the
  closed-form marginal posterior of any edge of the graph, given the
  correlation structure of the multivariate time series data on joint
  locations. On the basis of our recovery learning, we offer recommendations
  on the optimal exercise routines for patients with given level of mobility
  impairment.}
\end{abstract}

\noindent


\end{frontmatter}

\renewcommand\baselinestretch{1.0}
\section{Introduction}
\label{sec:intro}
\noindent
Medicine today demands early, patient-specific diagnosis. In this paper, we put forward a tool that paves the way for predicting the recovery trajectory of an individual patient to recover their lost mobility, as they undertake physical rehabilitation. Indeed, such mobility challenges can arise in a patient because of injuries or critical illnesses, such as strokes. In fact, stroke ranks as the second most common cause of death and the third-leading cause of adult disability, \cite{Feigin:2022}. Post-stroke symptoms - such as reduced mobility and discomfort caused by pain - are prevalent amongst stroke survivors. Here we introduce a new technique for reliably assessing the mobility recovery of patients with movement deficiencies, to construct each patient’s recovery trajectory, as they undergo physical rehabilitation. 

We learn the recovery of individual patients using a new distance/divergence
between random graph variables that are learnt using the time series datasets
generated as a patient repeats an exercise routine that is part of their
therapy. By analysing the constructed recovery trajectories, we can predict
the optimal exercise routines that are tailored to patients with varying
degrees of mobility impairment. Furthermore, as our identification of an
individual recovery trajectory is supplemented by the pre-therapy information
about the patient’s mobility status, we will in the future be able to
undertake the supervised learning of the relationship between the recovery
trajectory variable, and the patient’s condition at the beginning of the
therapy. This then will permit the all-important patient-specific and early
(i.e. pre-therapy) prediction of the recovery trajectory in a prospective
patient, in the context of a given therapeutic programme. Such prediction
demands the establishment of the currently-absent, but requisite training set,
i.e. a set of pairs of patient injury severity marker and their recovery
trajectories - this set is identified in this paper, by the learning of the
recovery trajectories.

There is a wide range of recovery quantification techniques reported in the
literature, and yet, a standardised definition of recovery is lacking. In 2001, the World Health Organization published the
International Classification of Functioning, Disability and Health (ICF)
\cite{WHO:2001}, as a model of classification relevant to healthcare-related
domains, and this has been widely used in studies of mobility recovery of
patients affected by movement-impeding conditions. This classification
technique addresses three main points associated with the influence
on health: body structure and function (both physical and mental); activity
limitation; and restriction in participation.
Indeed, ICF has worldwide usage as a reference model in the assessment of
functioning, such as in the cases following traumatic injuries and
stroke. The contribution of Walton (2009) \cite{Walton:2009} is also
important, as they
advanced 30 different primary operationalisations of recovery, (in the context
of acute Whiplash Associated Disorder or
WAD). Ritchie (2016) \cite{Ritchie:2016} summarised approaching the concept of
recovery from WAD using self-reporting of relevant symptoms, (such as pain and
dizziness); psychological observations such as pain catastrophising
 \cite{Casey:2015} and post-traumatic stress disorder symptoms. Kasch (2001) \cite{Kasch:2001}
and Ozegovi (2009) \cite{Ozegovic:2009} have considered recovery as
parametrised by proxy
measures such as reduction in working hours; sick leaves undertaken; and
insurance claims.

In contrast to the patient-specific recovery trajectories that we report here,
recovery trajectories that have been reported in the literature, refer to
grouping patients by recovery. These are often pursued using Growth Mixture
Modelling (GMM) and Latent Class Growth Analysis (LCGA) techniques
\cite{Nagin:1999, Muthen:2000,Nagin:2005}, applied to data comprising
patients' responses to administered questionnaires \cite{Walton:2016,
  Herrera:2007, Stark:2013}, or to patient data obtained from clinics
\cite{Royes:2010}. Jung and Wickrama (2008) \cite{Jung:2008} have considered
LCGA and GMM techniques to identify patient groups that are homogeneously
recovered; Ram and Grimm (2009) \cite{Ram:2009} summarised a four-step
procedure to implement a GMM-based analysis; Panken et al. (2016)
\cite{Panken:2016} undertake GMM to describe the evolution of the intensity of
lower-back pain; Lee et. al (2020) \cite{Lee:2020} identified meaningful
subgroups of patients who suffered from muscelo-skeletal trauma by using GMM;
Walton and colleagues (2021)
\cite{Walton:2021} have developed the Traumatic Injuries Distress Scale to
provide the magnitude and nature of risk for persistent problems in people
with muscelo-skeletal trauma, and have predicted the rate of recovery.

Our identification of the recovery trajectory of a patient is underlined by
the computation of a statistical distance/divergence between the posterior
probability of a random graph that we learn given a multivariate dataset -
generated by a patient undertaking an exercise routine - and the posterior of
another graph learnt given another dataset that is generated when the same
patient undertakes this routine at the subsequent instance. Then recovery
between the two successive undertakings of this exercise routine, is provided
by the inter-graph distance that we compute. Thus, this technique demands an
understanding of the random graphs that we work with here, within a Bayesian framework. We learn a random
graph variable given a multivariate dataset, so that we can express the
posterior probability of this random variable, conditional on said data. We
write the graph posterior as contributed to by the posterior probability of
each edge. In this way 
the graph learning here is distinct from graph learning undertaken elsewhere,
as in \cite{daitch, zhuzoubin}.

We
undertake the learning of a graph variable – given the relevant data - and
discuss the implementation of the inter-graph distance in
Section~\ref{sec:implementation}, In Section~\ref{sec:data generation} we
discuss the advantages of our method over existing approaches used for
assessing patient recovery. With the methodology discussed in
Section~\ref{sec:methodology}, in Section~\ref{sec:data} we present the data
that we use. Discussions on methodology are followed by a presentation of our
results in Section~\ref{sec:results}, We wrap up the paper with the
discussions of Section~\ref{sec:conclusion}.

\section{Data generation}
\label{sec:data generation}
A case is made for
at-home rehabilitation - enabled for example by telerehabilitation equipment that allows patients to
exercise in a home environment, unencumbered by the demand of travel to a
specialised rehabilitation centre, where such travel is particularly challenging for
low-mobility patients. At-home rehabilitation also offers the advantage of
avoiding long
waiting times and missed opportunities for
early - and tailored - rehabilitative interventions. In addition, research shows that
there is an increase over time in outpatient rehabilitation, as opposed to
inpatient rehabilitation \cite{Olasoji:2022}.

Here, by telerehabilitation, we refer to virtual reality-assisted rehabilitation,
which is a subtopic within the analysis of human motion. It is
facilitated by the use of sensors such as cameras and/or wearables. Exergames are a
particular type of virtual reality platforms that are successfully used in
physical rehabilitation as these exergames are designed to stimulate the
repetition of a certain movement, with a focus of improving physical
activity.
Machine learning-based assessment of individual movement correction has been undertaken in the past \cite{Osgouei:2020, Noureddin:2019, Capecci:2019, Paiement:2014}, with a comprehensive review on human motion quality assessment presented in \cite{Frangoudes:2022}. 
However, previous work suffer from several limitations: most of the
datasets are collected from healthy individuals rather than from patients;
typically, a universal benchmark is used, against which
improvement/deterioration is measured for all patients;
and that recovery is assessed at a time point rather than over a time interval. 
We offer a technique that will radically improve
physiotherapy delivery via an intelligent feedback based on an individual
patient's performance and needs.

Indeed, our method of learning individualised
recovery-trajectory can be integrated with telerehabilitation
equipment, informing the patient and their therapists of progress on recovery
made at any instance of the rehabilitation. The telerehabilitation
equipment extracts detailed information about movement biomarkers and 
collects such information using motion sensor devices. It is the time series
data on such biomarkers, generated while a patient is undertaking an exercise
routine, that we use towards the construct the ion of recovery trajectories.
The fast and
automated generation of a reliable mobility recovery score, learnt using such recorded
movement data, leads to the construction of individualised recovery trajectories. Such a construct is the first and the imperative step towards 
providing physiotherapist-free feedback to the
patient, (as well as the therapist), on their level of recovery.
We will learn such trajectories using a retrospective patient dataset, and advance
recommendations regarding optimal exercise routines for patients with a given
level of pre-therapy mobility deficiency. This is an important step in the attempt to automatically assess individualised physical functioning within a home environment.

\section{Methodology}
\label{sec:methodology}

Our focus is on the development of a tool aimed at providing patients and healthcare providers with accurate and comprehensive information regarding a patient's recovery, following engagement in a specific exergame for a predetermined number of sessions. It is imperative that
information extracted on patient recovery be accurate and robust; at the same time, such information needs
to be presented in a form that is easy to visualise and internalise. We
propose the utilisation of a reliably-learned patient recovery trajectory as
the desired tool for disseminating the sought recovery-related
information.

The primary challenge lies in establishing a trustworthy, explainable, and
automated learning of recovery trajectories, using the available data, namely,
the time series data comprising the recorded spatial coordinates of 20 joints
of the human body. These coordinates are recorded during the patient's
participation in the exergame sessions using a virtual e-platform called
\textit{MIRA Rehab} \cite{MIRA}. Spatial and rate coordinates of each of the
20 joints are recorded, as the $i$-th patient plays the $k$-th game, at the
$j$-th instance.
\begin{definition}
The total number of times the $i$-th patient plays the
$k$-th game is $N_{i,k}$; so $j=0,1,\ldots,N_{i,k}$, for $i=1,\ldots, N_p$ and
$k=1,\ldots, N_{g}$.
\end{definition}
\begin{remark}
In general, there may exist some $k$, such that
$N_{i,k}=0$ since the $i$-th patient will not necessarily play all the
available $N_g$ games.
\end{remark}
As a patient executes an exergame, the location of a joint in their body
changes with time - \textit{MIRA} tracks the location of 20 joints of the
patient's skeletal frame, throughout the execution of the exergame, at regular
intervals. This gives rise to a time series of patient
locations. A time-dependent location of a joint in the patient's body is
given - within a pre-fixed, 3-dimensional Cartesian coordinate frame - by a
3-dimensional vector of spatial coordinates. In other words, there are three
spatial coordinate variables that represent the location of each of the 20
joints, at a given time, during a patient's execution of a given exergame.

Time derivative of each spatial coordinate of a given joint is the
corresponding rate - that represents the rate of change of the location of the
joint - as the patient executes the exergame, and such rate variables are also
tracked by \textit{MIRA}. However, we do not use the rates in our work.

\begin{figure}
  {\hspace{1.5cm}
    \includegraphics[scale=0.5]{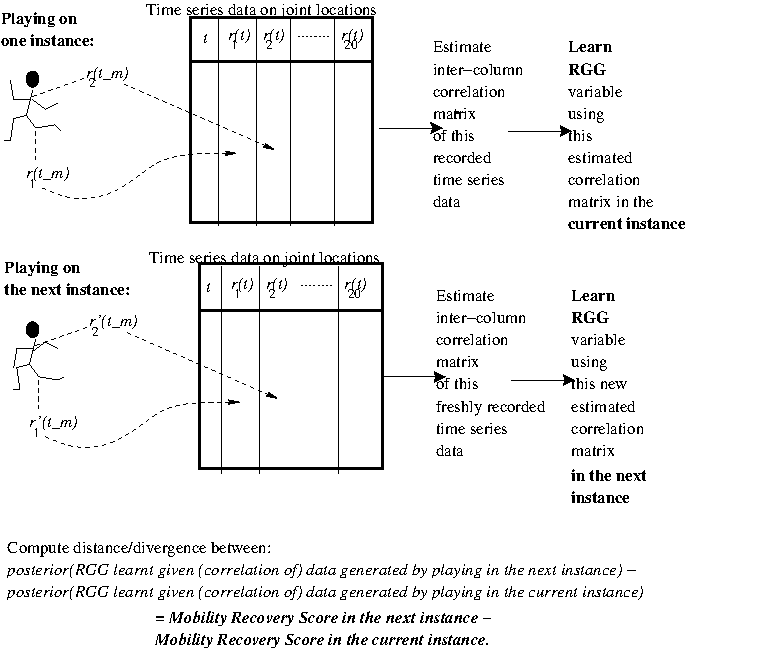}}
\caption{A cartoon of the basic framework of the method. The same patient
  plays a given exergame on successive instances - resulting in two
  successively recorded time series data on the locations of the 20 monitored
  joints in their body. The spatial coordinates of the three-dimensional
  location vector are periodically recorded on the e-platform {\textit{MIRA}}. The Euclidean norm of this vector then gives the location
  of a joint, at a time point when a recording is done by
  {\textit{MIRA}}. Thus, a time series (on joint location) data is produced as
  the patient plays this exergame. The inter-column correlation matrix of this
  data is estimated, and this estimated correlation used to learn a random
  graph variable - which is a Random Geometric Graph variable (or an RGG
  variable) in our
  work. Again, when this patient plays this exergame on the next instance, a
  new time series data is recorded; its inter-column correlation matrix
  estimated; and the RGG of this data is then learnt. A statistical
  distance/divergence measure between the posterior
  probabilities of the RGGs learnt given the (correlation of the) two datasets,
  is set proportional to the difference between the Mobility Recovery Scores
  attained (by this patient playing this exergame) at the next instance, and
  at the current instance.} 
\label{fig:cartoon}
\end{figure}

\begin{definition}
Let the three spatial coordinates of the $s$-th joint, recorded at time
$T=t$ during a subject playing a game, be denoted the variables $X_s(t),
Y_s(t),$  $Z_s(t)\in{\mathbb R}$, for $s=1,\ldots, 20$.
\end{definition}
A cartoon describing the relevance of the time series is displayed in
Figure~\ref{fig:cartoon}.

The strength of
the effect, i.e. the level of recovery of movement ability, is what needs to captured, while the $i$-th patient plays the $k$-th game in the
$j$-th instance, compared to when they play the game at the $j-1$-th
instance, $\forall j>1;\:\: j\leq N_{i,k}$.
\begin{remark}
In our work, we understand that the recovery of
movement will give rise to the differences between the ``correlation structure''
of the time series data recorded at the $j$-th instance of playing the game, and that
recorded at the previous instance of playing this game.
Here, the said correlation
structure comprises information on the interplay - or correlation - between each pair of the 20 joints of the human
body that are tracked for their location, as a subject undertakes an
exergame.
\end{remark}
After all, a patient who is facing movement difficulties in a
certain joint, will have to invoke usage of other (nearby) joints (called compensation), to perform
an action that involves the former joint. Basically, there exists higher correlation
amongst pairs of (nearby) joints in such a patient, than in the case for a
patient with less severe movement difficulties. Details of which joint
pairs will be more strongly correlated for a given patient, depend on the
nature of the movement difficulty; the undertaken task; and human anatomy. Thus, in a healthy patient who is not
suffering from movement difficulties, an action that calls for the usage of a
given joint, does not necessarily call for simultaneous usage of other joints,
i.e. the inter-joint correlations are then lower in general,
than in a patient with more severe mobility problems.

So we look at the correlation structure of locations of the monitored
joints as the marker of recovery of movement. The
correlation structure of a multivariate dataset is manifest in the graph of this
dataset. To contextualise this to our work, we will learn a graph of the time
series data that comprises values of spatial coordinates of each of 20 joints (monitored on the
e-platform {\textit{MIRA}}, on every instance when the $i$-th subject plays the $k$-th game), using the
inter-column correlation matrix of this multivariate time series data.

\subsection{Gist of the method used to construct recovery trajectories}
We now provide a gist of the method used to construct a recovery trajectory.
\begin{itemize}
\item In our work, we will treat the graph as a random variable, and learn this
random graph, given the correlation structure of the time series data
on locations of the 20 joints, as a patient plays an exergame.
\item Thus, there are 20 nodes in this random graph variable, with the
  location variable of a joint attached to a node. Given the inter-joint
  correlation of a recorded time series dataset, some nodes of the graph will be
  joined by a mutual edge, while the edge between some nodal-pairs will not exist
  in the learnt graph.
\item Since the graph is random, we can define a probability distribution of
  this random (graph) variable, conditional on the (correlation of the) time
  series data generated when the $i$-th patient plays the $k$-th game, in the
  $j$-th instance.
\item Then we can define a statistical distance/divergence between the
  probability of the graph variable learnt given the time series
  data generated when the $i$-th patient plays the $k$-th game in the $j$-th instance,
  and the graph learnt given the data recorded upon this patient playing the game in the
  $j-1$-th instance. This will be done for all instances relevant to the
  playing of the $k$-th game by the $i$-th subject, (i.e. $\forall j > 1;\:\:
  j \leq N_{i,k}$).
\item {\it{This distance computed for the $j, j-1$ graph pair is then a recovery
score obtained to inform on recovery that occurred in the $i$-th patient, between their playing of this game in the $j$-th instance, and the
previous instance.}} Such a recovery score can be both positive and negative in general. Using this recovery score, we can generate a recovery trajectory for the $i$-th
patient playing this $k$-th game. 
\end{itemize}
This learning of a recovery score is seated within a highly robust
mathematical technique - the implementation of which is transparent, and the
method
exploits all the available information. The constructed recovery trajectory
will be informative, and is a clear visual way of depicting the recovery that
the $i$-th patient has achieved, by playing the $k$-th game till the $j$-th
instance of playing.

\subsection{Model}
\label{sec:model}
\begin{definition}
We denote the location - at time $T=t$ - of the $s$-th joint (of the 20 joints tracked on the
e-platform by the {\it{MIRA}} facility) of a given patient's body, as they
play an exergame at a given instance, by the
Euclidean norm $R_s(t)$ of the $(X_s(t), Y_s(t), Z_s(t))^T$ vector. Thus, location of the $s$-th joint is the variable $$R_s(t) :=
\sqrt{(X_s(t))^2 + (Y_s(t))^2 + (Z_s(t))^2}.$$
\end{definition}
Then, as the $i$-th patient plays the $k$-th exergame on the $j$-th instance,
we use the multivariate time
series data on $X_1(t), Y_1(t), Z_1(t), X_2(t), Y_2(t), Z_2(t), \ldots,$
 $X_{20}(t), Y_{20}(t), Z_{20}(t)$, we
generate the time series data on $R_1(t), R_2(t), \ldots, R_{20}(t)$, i.e. we
record the time series of the locations of each of the 20 monitored joints.
\begin{definition}
In our work, we define ${\bf D}_{i,k,j}$ as the time series data that comprises values of
$R_1(t), \ldots, R_{20}(t)$, recorded on the
e-platform {\it{MIRA}} at a time point $t$, for $t=t_1, t_2, \ldots,
t_{i,k,j}^{(max)}$, as the  
$i$-th subject plays the $k$-th game, in the $j$-th instance.
\end{definition}
Thus, the considered time points of recording of $R_s(t)$
values is not continuous, i.e.
$T$ is a discrete variable for us.
We note that:
\begin{enumerate}
\item[---]the dataset ${\bf D}_{i,k,j}$ has
  $q_{i,k,j}^{(max)}$ rows and 20 columns, where $t_{i,k,j}^{(max)} =
  (q_{i,k,j}^{(max)} - 1)\epsilon_t + t_1$, with {\it{MIRA}} making observations of
  the 20 joints of the patient skeleton after every time interval of width
  $\epsilon_t$. Here $q_{i,k,j}^{(max)}\in{\mathbb N}$. Thus, ${\bf D}_{i,k,j}$ is a
  $q_{i,k,j}^{(max)}\times 20$-dimensional matrix. 
\item[---]Indeed, the length of time
needed by a subject to play a game in one instance can differ from the length
of time taken by the same or a different subject to play the same or a different game in
another instance.
\item[---]Thus, the time series data recorded for different instances of playing the same game, have a different temporal coverage even for the same patient.
\item[---]We need to compute the strength of the effect that is responsible for the dataset
${\bf D}_{i,k,j}$ to change into dataset ${\bf D}_{i,k,j^{/}}$, where the
temporal coverage of the two datasets is not necessarily the same,
i.e. $t_{i,k,j}^{(max)} \neq t_{i,k,j^{/}}^{(max)}$ in general, for $j\neq
j^{/}$; $j,j^{/}=1,\ldots, N_{i,k}$. This effect is caused by the recovery of
movement that the $i$-th patient has undergone, between them executing the
$k$-th game in the $j$-th instance and the $j^{/}$-th instance.
\end{enumerate}
\begin{definition}
As motivated above, we parametrise the
strength of the effect that causes ${\bf D}_{i,k,j}$ to change into dataset
${\bf D}_{i,k,j^{/}}$, by a distance/divergence between the probability of
the graph learnt given the inter-joint correlation matrix $\bSigma_{(i,k,j)}^{(C)}$ of the
dataset ${\bf D}_{i,k,j}$, and the graph learnt given the inter-joint
correlation matrix of the dataset ${\bf D}_{i,k,j-1}$. 
\end{definition}

\begin{definition}
We standardise the value of $R_s(t)$ using the mean and standard
deviation of the sample: $\{r_s^{(i,k,j)}(t_1), r_s^{(i,k,j)}(t_2), \ldots,
r_s^{(i,k,j)}(t_{i,k,j}^{(max)})\}$ that is recorded when the $i$-th patient plays the
$k$-th exergame on the $j$-th instance, with $r_s^{(i,k,j)}(t)$ denoted the
value attained by the location variable $R_s(t)$, at time $T=t$.
Here, the estimates of the sample mean ${\bar{r_s}}^{(i,k,j)}$ and sample standard deviation
$\zeta_{r_s}^{(i,k,j)}$ are given as:
$${\bar{r_s}}^{(i,k,j)} :=
\displaystyle{\frac{\sum\limits_{t=t_1}^{t_{i,k,j}^{(max)}}
    r_s^{(i,k,j)}(t)}{q_{i,k,j}^{(max)}}} 
\quad \zeta_{r_s}^{(i,k,j)} := \displaystyle{\sqrt{\frac{\sum\limits_{t=t_1}^{t_{i,k,j}^{(max)}}\left(r_s^{(i,k,j)}(t) - {\bar{r_s}}^{(i,k,j)}\right)^2}{(q_{i,k,j}^{(max)}-1)}}}.$$
We undertake such standardisation 
 $\forall s=1,\ldots, 20$. 
\end{definition}
\begin{remark}
The dataset generated by the standardised joint location values is
from now on, still referred to as ${\bf D}_{i,k,j}$, i.e. we do not introduce a new notation for the dataset comprising the standardised data. 
\end{remark}

Values attained by the random variable $R_s(t)$, comprise the $q$-th row and
the $s$-th column
of the dataset ${\bf D}_{i,k,j}$, (for $t=q\epsilon_t$). The fact that the number of rows of the dataset ${\bf
  D}_{i,k,j}$ and the dataset ${\bf D}_{i,k,j^{/}}$ are unequal, does not
affect our Bayesian learning of the
random graph variable that is realised given either data set,
and therefore, the inter-graph distance is
robust to the difference in the number of rows in either dataset.

\subsection{Learning/estimation of the correlation matrix}
\label{sec:correlation matrix}

\begin{definition}
Let the correlation matrix of dataset ${\bf D}_{i,k,j}$ be
$\bSigma_{i,k,j}^{(C)} = [\rho_{s,s^{/}}]$, where $\rho_{s,s^{/}}$ is
the correlation between the random variables $P_s^{(i,k,j)}$ and $P_{s^{/}}^{(i,k,j)}$, with the
$q$-th recorded (by {\textit{MIRA}}) value of $P_s^{(i,k,j)}$ realised at time
  $T=t=q\epsilon_t$, as
$r_s^{(i,k,j)}(t)$, when the $i$-th patient executes the $k$-th
game in the $j$-th instance. Here, $s,s^{/}=1,\ldots,
20$.
\end{definition}
Indeed, in our recorded data ${\bf D}_{i,k,j}$, realisation of
random variable $P_s^{(i,k,j)}$ recorded at $T=t$, concurs with the realisation of the variable
$R_s^{(i,k,j)}(t)$, $\forall t = t_1,\ldots,t_{i,k,j}^{(max)}$.

The following enumerates our addressing of the problem of learning/estimating
the correlation matrix $\bSigma_{i,k,j}^{(C)}$. 
\begin{enumerate}
\item[---]We could undertake the learning of each correlation element directly within
the inferential scheme that we could invoke, such as Markov Chain Monte Carlo
(or MCMC) based algorithms. However, there are $(20\times 20 -20)/2 = 190$
distinct off-diagonal elements of this symmetric correlation matrix, and
learning 190 (correlation) parameters using MCMC is a daunting - and in fact,
an impossibly daunting - task.
\item[---]It may appear possible to undertake
parametrisation of this correlation matrix using a covariance kernel, that
takes as its input a difference between the values of an input variable, and
produces the correlation between the outputs generated at these
inputs. However, there is no relevant variable that could serve as such an
``input" variable. The index of a joint in the human body - the location of
which is monitored at a given time point as the patient plays an exergame - is
merely nominal, and can be arbitrarily assigned to a joint, rendering the
difference between indices of two such joints bereft of any useful
interpretation.
\item[---]Then we cannot learn any element of the correlation matrix $\bSigma_{i,k,j}^{(C)}$; we can therefore only
estimate each element, using the sample of values of the location variable $P_s^{(i,k,j)}$ and that of $P_{s^{/}}^{(i,k,j)}$, as included across the 
$q_{i,k,j}^{(max)}$ rows of 
this time series dataset. We compute the estimate of the Pearson correlation coefficient.
\end{enumerate}

\begin{definition}
The unbiased estimator ${\hat{\rho}}_{s,s^{/}}$ of the correlation
$\rho_{s,s^{/}}$ between variables $P_s^{(i,k,j)}$ and $P_{s^{/}}^{(i,k,j)}$
as:
$${\hat{\rho}}_{s,s^{/}} =
\displaystyle{\frac{
    \displaystyle{\sum_{i=t_1}^{t_{i,k,j}^{(max)}}
      r_s^{(i,k,j)}(t) r_{s^{/}}^{(i,k,j)}(t) }}{q_{i,k,j}^{(max)}-1}},$$
recalling that the data on $P_s^{(i,k,j)}$ is standardised. This holds for
$s, s^{/}=1, \ldots, 19$.
\end{definition}
Then the inter-column correlation matrix estimated in data ${\bf D}^{(i,k,j)}$
is ${\hat{\bSigma}}_{i,k,j}^{(C)} = [{\hat{\rho}}_{s,s^{/}}]$.

\begin{definition}
The partial correlation 
$\psi_{s,s^{/}}$ between variables $P_s^{(i,k,j)}$ and $P_{s^{/}}^{(i,k,j)}$
is computed using the estimated $\theta_{s,s^{/}}$, given the
precision matrix $\bTheta_{i,k,j} = [\theta_{s,s^{/}}] \coloneqq
(\bSigma_{i,k,j}^{(C)})^{-1}$. In fact, the inter-column partial correlation
matrix of dataset ${\bf D}_{i,k,j}$
is $\bPsi_{i,k,j}^{(C)} =[\psi_{s,s^{/}}]$, for $s,s^{/}=1.\ldots,20$, where
\begin{equation}\label{eq:1}
    \psi_{s,s^{/}} = -\frac{\theta_{s,s^{/}}}{\sqrt{\theta_{s,s}\theta_{s^{/},s^{/}}}},
    s \neq s^{/},\text{ and we have }\psi_{s,s} = 1\text{ for }s =  s^{/}.
\end{equation}
\end{definition}


\subsection{Outlining graph learning $\&$ inter-graph distance computation}
\begin{enumerate}
\item Once the partial correlation 
$\psi_{s,s^{/}}$ between variables $P_s^{(i,k,j)}$ and $P_{s^{/}}^{(i,k,j)}$
is computed, we learn the value of the (binary) edge $G_{s,s^{/}}$ variable
that joins the 
$s$-th and $s^{/}$-th nodes, (i.e. learn if $g_{s,s^{/}}$ is 1, or is instead
0), given $\psi_{s,s^{/}}$. We recall that random variables $P_s^{(i,k,j)}$
and $P_{s^{/}}^{(i,k,j)}$ sit at the $s$-th and $s^{/}$-th nodes
respectively. In our Bayesian setting, we perform this edge
learning by computing the posterior probability of this edge, conditional on
$\psi_{s,s^{/}}$, for each relevant $s,s^{/}$ pair, i.e. $\forall s,s^{/}; \: s < s^{/}; \:
s=1,\ldots,19$.
\item Thus,
the edge between the $s$-th and $s^{/}$-th nodes is modelled to be affected by
the partial correlation between {\it{only}} the variables $P_s^{(i,k,j)}$ and
$P_{s^{/}}^{(i,k,j)}$, while values of all other variables
are held as fixed. Another possibility would be to learn $G_{s,s^{/}}$ as conditional on
the correlation between $P_s^{(i,k,j)}$ and
$P_{s^{/}}^{(i,k,j)}$, computing which does not demand that all other
variables be held fixed.
\item In this work, we write the edge
probability of $G_{s,s^{/}}$ conditional on the data - namely, the partial correlation
  $\psi_{s,s^{/}}$ between the nodes that this edge joins - as
$m(G_{s,s^{/}}\vert \psi_{s,s^{/}})$. We will identify a closed-form
expression for this conditional
probability ($m(G_{s,s^{/}}\vert \psi_{s,s^{/}})$) below (in Equation~\ref{eqn:2}).
We will include the edge between the $s$-th and $s^{/}$-th nodes, only if the edge probability exceeds a
chosen ``cut-off'' probability $\tau\in[0,1]$, i.e. $g_{s,s^{/}}=1$,
if $m(G_{s,s^{/}}=g_{s,s^{/}}\vert \psi_{s,s^{/}})\geq \tau$.
\item Now, to check if $m(G_{s,s^{/}}\vert \psi_{s,s^{/}})\geq \tau$, we need
  to compute 
$m(G_{s,s^{/}}=g_{s,s^{/}}\vert \psi_{s,s^{/}})$, and to compute this
  posterior probability, we need to know the value
$g_{s,s^{/}}$ of the edge variable $G_{s,s^{/}}$. 
We recognise the problem of learning $g_{s,s^{/}}$ from the posterior
  probability of $G_{s,s^{/}}$ given the data, as a problem in
  Bayesian inference. We sample $g_{s,s^{/}}$ from this edge
  posterior, the closed-form expression of which is available, as we have
  stated above. We undertake Rejection Sampling to sample from the edge posterior
$m(G_{s,s^{/}}=g_{s,s^{/}}\vert  \psi_{s,s^{/}})$, for all relevant nodal
pairs. As in any implementation of Bayesian inference, we will then
  have $N$ sampled $g_{s,s^{/}}$ values, and the edge posterior at each
  sampled edge value.
  \begin{enumerate}
  \item[---]The set of posterior values for all relevant edge
  variabes will define the posterior of the graph variable, given the partial
  correlation matrix of one dataset, 
  and we will compute the inter-graph distance/divergence between the graph
  posterior given this dataset and that given another, as per
Definition~\ref{defn:hell}/\ref{defn:kl}.
  \item[---]The set of sampled edge values will be employed to compute
  an estimate of the edge posterior probability, which if $\geq \tau$, 
  $G_{s,s^{/}}$ is set to 1 in a realisation of the graph variable at a chosen
  $\tau$; else we set $G_{s,s^{/}}$ to 0. This is discussed
in Definition~\ref{defn:prothom}.
\end{enumerate}
\item One question remains unanswered still: what is the justification behind using
$m(G_{s,s^{/}}=g_{s,s^{/}}\vert  \psi_{s,s^{/}})\geq \tau$ to check for the
existence of an edge? The answer is that this inequality follows from our
modelling of the random graph as a
Random Geometric Graph, (or RGG), any edge of which is a probability. Such an RGG is manifest if we draw the RGG in a
probablistic metric space. We discuss such an RGG next.
\end{enumerate}

\subsection{Random graph variable}
\label{sec:graph variable}
The graph of the $q_{i,k,j}^{(max)}\times 20$-dimensional
dataset ${\bf D}_{i,k,j}$, is defined on the vertex set
$\bV=\{P^{(i,k,j)}_1,\ldots, P^{(i,k,j)}_{20}\}$. There are no self-loops
allowed in this graph and the edges form independently of each other. This graph is drawn
using the $20\times 20$-dimensional partial correlation matrix
$\bPsi_{i,k,j}^{(C)}$, any element of which can be computed by
Equation~\ref{eq:1}. As stated above, in this work, the random graph variable is a Random
Geometric Graph \cite{gilbert_61, giles, Penrose:2016, plos} drawn in a probabilistic metric space \cite{sklar}.
\begin{definition}
A Random Geometric Graph (RGG) is s.t. an edge exists between a pair of nodes,
if and only if, the distance between the nodes falls below a pre-fixed cutoff
or threshold $\tau$.  
\end{definition}
Therefore, in an RGG, the affinity function between the pair of variables that are
attached to a pair of nodes - where the affinity is complimentary to the
inter-nodal distance function - exceeds or equals $\tau$ to ensure that the
mutual edge exists; else the edge is absent.

\begin{definition}
In a probabilistic metric space, to any pair of points in this space, we can
assign a probability distribution of a non-negative function of this pair of
points, s.t. the distribution is over non-negative support. 
\end{definition}
Just as a non-negative distance can be assigned to a pair
of points in a metric space, in a probabilistic metric space, a probability distribution with
non-negative support is assigned to a pair of points\footnotemark.
\footnotetext{Formally, a
probabilistic metric space is a triple comprising: the sample space that hosts
random variables; a probabilistic distance function between any two points in
this space; and a triangle function that underwrites this distance function by
assuring its adherence to the triangle rule. Said probabilistic distance
function is a
probability distribution over non-negative support.}

Thus, any edge of the RGG variable drawn in probabilistic metric space, is a
probability. As the (probabilistic) distance between variables $P_s^{(i,k,j)}$
and $P_{s^{/}}^{(i,k,j)}$ increases, there is a decrease in the probability
that an edge will join the nodes that these variables sit at. In other words,
as the value of the (complimentary) affinity function between $P_s^{(i,k,j)}$
and $P_{s^{/}}^{(i,k,j)}$ decreases, there is an decrease in the edge
probability. Similarly, increasing affinity implies edge probability is
increasing. The
affinity is then intuited to be given by the probability of the edge variable
$G_{s,s^{/}}$. Then as in an RGG, we demand that only if
the affinity between $P_s^{(i,k,j)}$ and $P_{s^{/}}^{(i,k,j)}$ exceeds a
chosen cutoff $\tau$ - i.e. only if edge probability exceeds $\tau$ - the edge exists between the $s$-th and $s^{/}$-th nodes.

\subsection{Edge posterior $\&$ samples generated from it}
Here, this probabilistic distance between any two nodes of the RGG is a
measure of the {\it{intra-graph}} distance function, while our interest in
this application is in the distance/divergence between a pair of RGGs learnt
given two distinct time series (on joint locations) datasets, where the
{\it{inter-graph}} distance/divergence is defined between posterior
probability of a graph variable learnt given one dataset, and that learnt
given another data. The posterior of the graph is in turn contributed to by
the posterior probability of each edge variable that connects each nodal pair
of this graph.

We now enumerate steps to the achievement of this edge posterior.
\begin{enumerate}
\item[---]As motivated by \cite{chakrabarty:2023}, the first step in the formulation of the edge
  marginal is to model
the probability density function of the partial correlation $\psi_{s,s^{/}}$, as ${\cal
  N}(g_{s,s^{/}}, \upsilon_{s,s^{/}})$, where $\upsilon_{s,s^{/}}$ is the
variance parameter relevant to this ($s,s^{/}$-th) nodal pair.
\item[---]Then, in Bayes rule, we input this density of the (observable) partial correlation
  conditional on the edge and variance parameter, and 
use weak priors on the two model parameters - Bernoulli(0.5) on $G_{s,s^{/}}$ and
Uniform on $\upsilon_{s,s^{/}}$ - to write the joint posterior probability
density of $G_{s,s^{/}}$ and $\upsilon_{s,s^{/}}$, given partial
correlation $\psi_{s,s^{/}}$.
\item[---]This joint posterior is then proportional to: $(2\pi\upsilon_{s,s^{/}})^{-1/2}
  \exp(-(g_{s,s^{/}} - \psi_{s,s^{/}})^2/2\upsilon_{s,s^{/}})$, since the priors
  are constants with respect to the parameters.
\item[---]We then marginalise the variance parameter out of this joint
  posterior by integrating this joint over values of $\upsilon_{s,s^{/}}$ in
  the interval $(0, u]$ for a chosen $u>0$. We provide one model for the edge
    marginal posterior probability given the partial correlation between
    $P_s^{(i,k,j)}$ and $P_{s^{/}}^{(i,k,j)}$ by choosing $u=1$, where $(0,1]$
      is shown as an exhaustive set for $\upsilon_{s,s^{/}}$ to take values
      in.
\item[---] The resulting integral - i.e. the edge marginal posterior - is given as:
\begin{eqnarray}\label{eqn:2}
    &m(G_{s,s^{/}}\vert \psi_{s,s^{/}}) = 
    K \Bigg[\sqrt{\frac{2}{\pi}} \exp{\Bigg(\frac{-(S_{s,s^{/}})^2}{2}}\Bigg)
    - S_{s,s^{/}}\text{erfc} \Bigg (\frac{S_{s,s^{/}}}{\sqrt{2}}\Bigg)
    \Bigg],& \nonumber \\
&{\text{where }} S_{s,s^{/}} \coloneqq |G_{s,s^{/}} - |\psi_{s,s^{/}}|| \in
    [0,1].& \nonumber \\   
\end{eqnarray}
Here
the complementary error function $\text{erfc}(\cdot) = 1-
\text{erf}(\cdot)$, and $K > 0$ is an identifiable constant.
\end{enumerate}

\begin{definition}
Modelling edges to form independently of each other,
the posterior probability of the random graph variable ${\cal
  G}_{i,k,j}(\bV)$, conditional on the partial correlation of data ${\bf
  D}_{i,k,j}$, is the product over all relevant $s,s^{/}$ pairs, of the
conditional edge marginal given in Equation~\ref{eqn:2}:
\begin{eqnarray}\label{eq:3}
    &\pi({\cal G}_{i,k,j}(\bV)\vert {\bf D}_{i,k,j}) = \prod^{N-1}_{s <
    s^{/}; s =1} m(G_{s,s^{/}}\vert \psi_{s,s^{/}})& \nonumber \\
    &= K^{/} \prod^{N-1}_{s < s^{/}; s =1} \Bigg[\sqrt{\frac{2}{\pi}} \exp{\Bigg(\frac{-(S_{s,s^{/}})^2}{2}}\Bigg)
    - S_{s,s^{/}}{\textrm{erfc}} \Bigg (\frac{S_{s,s^{/}}}{\sqrt{2}}\Bigg) \Bigg]&,
\end{eqnarray}
\end{definition}
It is from the closed-form marginal posterior probability
$m(G_{s,s^{/}}\vert \psi_{s,s^{/}})$ (of edge $G_{s,s^{/}}$) given in
Equation~\ref{eqn:2}, that we generate $N$ samples of $G_{s,s^{/}}$ using
Rejection Sampling. This sample is:
$$\xi_{s,s^{/}} := \{g_{s,s^{/}}^{(1)}, \ldots, g_{s,s^{/}}^{(N)}\}.$$ Here, $g_{s,s^{/}}^{(r)}$ can be 1 or 0,
$\forall r=1,\ldots,N$. This is undertaken $\forall s,s^{/}; \: s < s^{/}; \:
s=1,\ldots,19$. Details of this undertaken Rejection sampling are
given in Section~\ref{sec:rejection sampling}.

Posterior of the graph variable
${\cal G}_{i,k,j}(\bV)$ constructed using the $r$-th sampled value of
$G_{s,s^{/}}$, for all relevant $s,s^{/}$-th pairs - given data ${\bf
  D}_{i,k,j}$ - is denoted $\pi({\cal G}_{i,k,j}(\bV)\vert {\bf
  D}_{i,k,j})^{(r)}$, where $r=1,2,\ldots,N$. We use $\pi({\cal
  G}_{i,k,j}(\bV)\vert {\bf D}_{i,k,j})^{(1)}, \ldots,$ $\pi({\cal
  G}_{i,k,j}(\bV)\vert {\bf D}_{i,k,j})^{(N)}$ for the computation of the
inter-graph distance/divergence, (discussed in Section~\ref{sec:inter-graph distance}). 

\subsection{Realisation of the graph variable at a chosen $\tau$}
While we will compute the inter-graph distance/divergence, we will also
visualise the graph generated using Rejection Sampling, at a given choice of
$\tau$ - constructed using Definition~\ref{defn:prothom} (that invokes Definition~\ref{defn:nu}).
\begin{definition}
\label{defn:nu}  
The relative frequency of edge variable $G_{s,s^{/}}$ in the
sample $\xi_{s,s^{/}}$ is $\nu_{s,s^{/}} = {{\sum_{n=1}^N
    g_{s,s^{/}}^{(n)}}/{N}.}$

\end{definition}
Then $\nu_{s,s^{/}}$ estimates $m(G_{s,s^{/}}\vert \psi_{s,s^{/}})$ in the
sample $\{g_{s,s^{/}}^{(1)}, g_{s,s^{/}}^{(2)}, \ldots, g_{s,s^{/}}^{(N)}\}$.

\begin{definition}
\label{defn:prothom}  
In our work, the realisation ${\cal G}_{\bV, m}(\bPsi,
\tau)$ of the graph variable is learnt at the cutoff $\tau$,
\begin{enumerate}
\item[---]on
the vertex set $\bV$; using the edge probability $m(\cdot\vert\cdot)$ defined
in Equation~\ref{eqn:2}; for the inter-column partial correlation matrix
$\bPsi_{i,k,j}^{(C)}$ of the dataset ${\bf D}_{i,k,j}$; and for the chosen cut-off $\tau$,
\item[---]where edge
between the $s$-th and $s^{/}$-th nodes exists, (i.e. $g_{s,s^{/}}=1$) if and
only if $\nu_{s,s^{/}} \geq \tau$, $\forall s,s^{/}; \: s < s^{/}; \:
s=1,\ldots,19$. Here, $\nu_{s,s^{/}}$ is defined in Definition~\ref{defn:prothom}.
\end{enumerate}
\end{definition}




\subsection{Inter-graph distance formalisation}
\label{sec:inter-graph distance}
A statistical distance or divergence can be computed between the posterior probability of
the graph variable conditional on the time series data ${\bf D}_{i,k,j}$, and
that given data ${\bf D}_{i,k,j^{/}}$. The Hellinger distance is an example of
such a statistical distance while the Kullbeck Leibler divergence can also be
computed.
\begin{definition}
The squared Hellinger
distance between the probability density $p_1(x)$ and another density $p_2(x)$
(for $x\in {\cal X}$) is:
\begin{equation*}
D^2_H(p_1,p_2) = \int_{\cal X} (\sqrt{p_1(x)} - \sqrt{p_2(x)})^2 dx.
\end{equation*}
\end{definition}

\begin{definition}
  \label{defn:hell}
Recalling that the graph posterior probability is computed at each of the $N$
number of samples generated using Rejection Sampling, for the random graph variables learnt given data ${\bf D}_{i,k,j}$ and ${\bf
  D}_{i,k,j^{/}}$, the squared discretised Hellinger distance between their
posterior probabilities 
is
\begin{eqnarray}
  \label{eq:4}
    & D^2_H({\cal G}_{i,k,j}(\bV), {\cal G}_{i,k,j^{/}}(\bV)) =&
  \nonumber \\
    &\displaystyle{\frac{\sum\limits^N_{r=1}\Bigg(\sqrt{\pi({\cal G}_{i,k,j}(\bV)\vert {\bf D}_{i,k,j})^{(r)}} - \sqrt{\pi({\cal G}_{i,k,j^{/}}(\bV)\vert {\bf D}_{i,k,j^{/}})^{(r)}}\Bigg)^2}{N}},&
\end{eqnarray}
where ${\pi({\cal G}_{i,k,j}(\bV)\vert {\bf D}_{i,k,j})^{(r)}}$ is the graph posterior given the data ${\bf D}_{i,k,j}$, computed with the $r$-th sample of the edges of the graph, $\forall r=1,\ldots,N$.  
\end{definition}

We compute the Hellinger distance $D_H({\cal G}_{i,k,j}(\bV),{\cal
  G}_{i,k,j-1}(\bV))$ between graph posterior given the data ${\bf D}_{i,k,j}$
and that given data ${\bf D}_{i,k,j-1}$, where these datasets are generated by the $i$-th patient playing the
$k$-th exergame, in the $j$-th and $j-1$-th instances, for all $i$ and $k$ relevant in our data to $j>1$, i.e. for all
patients who play an exergame multiple times.

\begin{definition}
\label{defn:kl}
Alternatively, we can compute the Kullbeck-Leibler divergence between the
posteriors of the graphs ${\cal G}_{i,k,j}(\bV)$ and ${\cal
  G}_{i,k,j-1}(\bV)$, given the data ${\bf D}_{i,k,j}$ and ${\bf
  D}_{i,k,j-1}$, as:
\begin{eqnarray}
  \label{eq:5}
    & D_{KL}({\cal G}_{i,k,j}(\bV), ({\cal G}_{i,k,j-1}(\bV)) =&
  \nonumber \\
    &\displaystyle{\sum\limits_{r=1}^N \pi({\cal G}_{i,k,j}(\bV)\vert {\bf D}_{i,k,j})^{(r)} \text{log} \Bigg (\frac{\pi({\cal G}_{i,k,j}(\bV)\vert {\bf D}_{i,k,j})^{(r)}}{\pi({\cal G}_{i,k,j-1}(\bV,\tau)\vert {\bf D}_{i,k,j-1})^{(r)}} \Bigg)},&
\end{eqnarray}
where the posterior of the random graph variable is computed given the datasets
generated by the $i$-th subject playing the $k$-th game on successive
instances, i.e. at the $j-1$-th and $j$-th instances, for $j =2, \ldots
N_{i,k}^{(max)}$.
\end{definition}

The kind of random graphs that we have learnt in our
current work, could be considered to fall within the broad class of
Inhomogeneous Random Graphs \cite{remco, node}. 

\subsection{Construction of recovery trajectories}
\label{sec:recovery trajectories MRS}
The posterior probability of a random graph variable - given 
time-series datasets on joint locations generated by a patient playing a given
exergame at successive instances - will be utilised to compute the inter-graph
distance/divergence between adjacent instances. Such an inter-graph distance computed
between the $j$-th and the $j-1$-th instances of the $i$-th patient playing
the $k$-th game will inform on the movement recovery attained by this patient
between these two instances of playing the game. (Here $j=2,\ldots,
N_{i,k}^{(max)}$).
\begin{definition}
Movement recovery is treated as the difference between the “Mobility Recovery
Score” or MRS ${\cal MRS}_{i,k}(j)$ attained by the $i$-th patient when
playing the $k$-th exergame in the $j$-th instance, and the score ${\cal
  MRS}_{i,k}(j-1)$ attained in the $j-1$-th instance of playing this game. We
model the MRS as proportional to the inter-graph distance/divergence between
the random graph variables learnt given datasets ${\bf D}_{i,k,j}$ and ${\bf
  D}_{i,k,j-1}$. We define the constant of proportionality to be unity, since we are defining the MRS
here.
\end{definition}

When we compute the MRS using the Hellinger distance, we denote the score as
${\cal MRS}^{(Hell)}_{i,k}(\cdot)$. On the other hand, when we use the KL-divergence to
give the value of the MRS, we denote the score ${\cal MRS}^{(KL)}_{i,k}(\cdot)$. We can
model the difference between ${\cal MRS}^{(Hell)}_{i,k}(j)$ and ${\cal MRS}^{(Hell)}_{i,k}(j-1)$
as proportional to the Hellinger distance between the graph variables learnt
given the time series data ${\bf D}_{i,k,j}$ and ${\bf D}_{i,k,j-1}$
respectively, i.e.
$${\cal MRS}^{(Hell)}_{i,k}(j) - {\cal MRS}^{(Hell)}_{i,k}(j-1) \propto D_H({\cal
  G}_{i,k,j}(\bV), {\cal G}_{i,k,j^{/}}(\bV)),$$ where we advance
  the value of 1 for the proportionality constant, as stated above.

\begin{remark}
Now, ${\cal MRS}_{i,k}(j)-{\cal MRS}_{i,k}(j-1)$ is the difference in the MRS between the
$j$-th and $j-1$-th instances of playing the $k$-th game. So this difference
is the same as the rate of change of MRS with respect
to instance of playing, at the $j-1$-th instance of playing
this game. Thus, the $i$-th patient$^{,}$s rate of change of mobility at
the $j$-th instance of playing the $k$-th game, using Hellinger distance is $
D_H({\cal G}_{i,k,j}(\bV), {\cal G}_{i,k,j^{/}}(\bV)),$ such that
their improvement score - or MRS - at this $j$-th instance of playing this
game is $$ {\cal MRS}^{(Hell)}_{i,k}(j) = D_H({\cal G}_{i,k,j}(\bV), {\cal
  G}_{i,k,j^{/}}(\bV)) + {\cal MRS}^{(Hell)}_{i,k}(j-1),$$ where $j=2,\ldots, N_{i,k}^{(max)}$.
\end{remark}

Again, when using the KL-divergence to compute the
inter-graph distance, we set the improvement score of the $i$-th patient at
the $j$-th instance of playing the $k$-th game to be:
$${\cal MRS}^{(KL)}_{i,k}(j) = D_{KL}({\cal G}_{i,k,j}(\bV), {\cal
  G}_{i,k,j^{/}}(\bV)) + {\cal MRS}^{(KL)}_{i,k}(j-1),$$
with $j=2,\ldots,N_{i,k}^{(max)}$.

We set ${\cal MRS}^{(\cdot)}_{i,k}(1)=0$ and plot ${\cal MRS}^{(\cdot)}_{i,k}(j)$ against $j$.
\begin{definition}
The variation of ${\cal MRS}^{(\cdot)}_{i,k}(j)$ with $j$ constitutes the
recovery trajectory for the $i$-th patient, as they play the $k$-th game.
\end{definition}

Since the posterior of a random graph variable, given the time series data on the 20
joint locations is the joint posterior of all edges of the RGG with
(20$^2$ - 20)/2=190 nodes, the posterior of the graph variable ${\cal
  G}_{i,k,j}(\bV)$ is a small number. Therefore, the distance/divergence
between the posterior of random graph variable ${\cal G}_{i,k,j}(\bV)$ given data
${\bf D}_{i,k,j}$, and of ${\cal G}_{i,k,j-1}(\bV)$ given data ${\bf
  D}_{i,k,j-1}$, is typically a small number. In fact, the divergence values
are smaller than the distance values. To avoid working with such small
numbers, we scale the posterior of any random graph variable, learnt using a
dataset that we consider in our work, by $10^{15}$, when computing Hellinger
distance between any pair of graph posteriors, given any pair of time series
datasets. Again, when computing the Kullbeck Leibler divergence, we scale any
such posterior value with $10^{25}$.

We have confirmed the robustness of the undertaken Rejection sampling, to variation in the proposal density. In Section~\ref{sec:robustness}, we demonstrate recovery trajectories that are learnt using a Bernoulli proposal, as well as a Uniform proposal. 

\begin{algorithm}[!t]
 \SetAlgoLined
 \SetKw{KwBy}{increment by}
 \SetKwInOut{Input}{Input}
 \SetKwInOut{Output}{Output}
$\forall j=2,\ldots,N_{i,k}^{(max)}$). \\
\tcc{Start learning recovery trajectory for $i$-th patient playing the $k$-th
  exergame, on the $j$-th instance}
\text{Spatial coordinate values $x_s^{(i,k,j)}(t), y_s^{(i,k,j)}(t),
  z_s^{(i,k,j)}(t)$ of the $s$-th joint}\\
\text{recorded at $t=t_1,t_2,\ldots, t_{i,k,j}^{(max)}$, by
    {\it{MIRA}}, as this exergame is played; $s=1,\ldots,20$.}\\
  \For{$t\gets1$ \KwTo $t_{i,k,j}^{(max)}$ \KwBy $1$,}{
    \For{$s\gets1$ \KwTo $20$ \KwBy $1$,}{
  \text{Compute location value $r_s^{(i,k,j)}(t)$ of the $s$-th joint at time
    $T=t$, as:}\\
  \text{$r_s^{(i,k,j)}(t) =
  \sqrt{(x_s^{(i,k,j)}(t))^2 + (y_s^{(i,k,j)}(t))^2 + (z_s^{(i,k,j)}(t))^2}$.} 
  }}
  \text{Gives rise to $t_{i,k,j}^{(max)}\times 20$-dimensional time series (on joint location) dataset.}\\
  \text{Estimate $20\times 20$-dimensional inter-column correlation matrix
    $\bSigma_{i,k,j}^{(C)}$ as per Definition~3.8.}\\
  \text{Compute partial correlation matrix $\bPsi_{i,k,j}^{(C)}$ using
    $\bSigma_{i,k,j}^{(C)}$ as per Definition~3.9.}\\
  \For{$r\gets1$ \KwTo $N=50,000$ \KwBy $1$,}{
    \For{$s\gets1$ \KwTo $19$ \KwBy $1$,}{
        \For{$s^{/}\gets s+1$ \KwTo $20$ \KwBy $1$,}{
            \text{Given $s,s^{/}$-th element of $\bPsi_{i,k,j}^{(C)}$, sample
              $g_{s,s^{/}}$ from edge marginal}\\
            \text{$m(G_{s,s^{/}}\vert \bPsi_{i,k,j}^{(C)})$ given in Equation~3.2}\\
        }
        }
  \text{Compute posterior probability $\pi({\cal
      G}_{i,k,j}(\bV)\vert\bPsi_{i,k,j}^{(C)})^{(r)}$
    of random graph variable}\\
  \text{${\cal
      G}_{i,k,j}(\bV$ given $\bPsi_{i,k,j}^{(C)}$. at the $r$-th sample
    of edges $\{g_{s,s^{/}}^{(r)}\}_{s^{/} > s; s^{/}=2}^{20}$.}\\
  }
\text{Use $\{\pi({\cal
      G}_{i,k,j}(\bV)\vert\bPsi_{i,k,j}^{(C)})^{(r)}\}_{r=1}^{50,000}$ to compute $D_H({\cal G}_{i,k,j}(\bV), {\cal G}_{i,k,j-1}(\bV))$}\\
\text{using Equation~3.4, and $D_{KL}({\cal G}_{i,k,j}(\bV), {\cal G}_{i,k,j-1}(\bV))$ using
  Equation~3.5.}\\
\text{Compute ${\cal MRS}^{(Hell)}_{i,k}(j) = D_H({\cal G}_{i,k,j}(\bV),
  {\cal G}_{i,k,j-1}(\bV)) + {\cal MRS}^{(Hell)}_{i,k}(j-1)$}\\
\text{and ${\cal MRS}^{(KL)}_{i,k}(j) = D_{KL}({\cal G}_{i,k,j}(\bV),
  {\cal G}_{i,k,j-1}(\bV)) + {\cal MRS}^{(KL)}_{i,k}(j-1)$. Here}\\
\text{${\cal MRS}_{\cdot}^{(\cdot)}(1):=0$. Record $(j, {\cal MRS}^{(Hell)}_{i,k}(j))$, and $(j, {\cal
    MRS}^{(KL)}_{i,k}(j)$.}\\ 

\tcc{Recovery trajectory using Hellinger distance function is plotted using $\{(j,
{\cal MRS}^{(Hell)}_{i,k}(j))\}_{j=2}^{N_{i,k}^{(max)}}$ and using KL divergence is
plotted using $\{(j, {\cal MRS}^{(KL)}_{i,k}(j))\}_{j=2}^{N_{i,k}^{(max)}}$.}
\caption{Algorithm for the computation of recovery trajectory for the $i$-th patient
  who plays the $k$-th exergame, on $N_{i,k}^{(max)}$ successive instances.}
\label{algo:1st}
\end{algorithm}

\section{Data}
\label{sec:data}
\begin{figure}[!h]
{\includegraphics[scale=0.35]{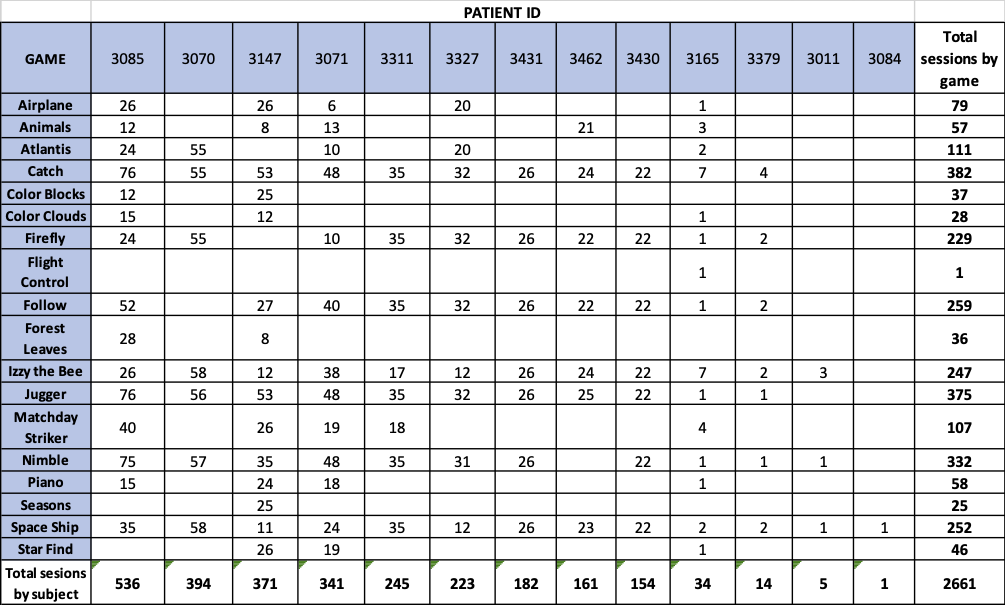}}
\caption{Sessions played by each subject for each of the 18 games.}
\label{figure1}
\end{figure}

The data was collected while a cohort of 
patients - who were suffering mobility deficiencies due to a stroke and were
undergoing physical rehabilitation - played exergames on the {\textit{MIRA}}
platform. This is a virtual reality platform and the users interact with the
exergames via a Kinect motion sensor camera. Data consists of recordings of
the three spatial coordinates ($X$, $Y$, $Z$) of 20 joints of the
skeletal structure of the patient who is playing an exergame. An exergame
engages the subject in the execution of a certain physical exercise or a
combination of two or more exercises (for instance: knee flexion; knee
extension; lateral flexion; shoulder abduction; frontal flexion; hip
abduction; full body turn, etc).  The execution of any exergame is captured as
a sequence of frames, with each frame capturing the three spatial coordinates
of each of the 20 considered joints. This translates into a sequence over the
time taken to play the exergame, or a time series - in each dimension of each
of the 20 considered joints. In our work, we reduce the dimensionality of this
recorded time series from 60 to 20, by agglomerating the three spatial
coordinates to the location $R :=\sqrt{X^2+Y^2+Z^2}$. Typically, a
patient$^{,}$s movement is recorded for 3 minutes at a frequency of
30Hz. In the cohort, 13 patients have played various games, with multiple
repetitions of each exergame, and at various levels of difficulty, as
presented in Figure \ref{figure1}. Overall, 2,661 sessions were recorded, and
these are not uniformly distributed amongst patients and exergames.

Table~\ref{table:1} presents a snippet of the data available for an individual patient in the cohort.
This example patient is one with reference 3071 in our
database. The table contains the time series data on the 20 joint locations while the patient
played an exergame called \textit{Airplane} that involves physical exercises such as elbow flexion; movement of the arm; moving the shoulder; moving the spine and moving the hip. In Table~\ref{table:1}, we present the first few lines of the
values of locations of some of the 20 joints in the patient's body, as
monitored on the e-platform {\textit{MIRA}}, as this patient (with identity
reference ID3071) played \textit{Airplane}.

The e-platform outputs several statistics at the end of each session (or
undertaken exergame), such as moving time in exercise; still time; average
correct answer reaction time; average incorrect answer reaction time; etc., as
well as the points gained for each exergame played. These inbuilt points are
expected to function as an indicator of patient performance in executing the
exergame - except, as we will notice in the results, these points do not offer meaningful information. 
\begin{table}[H]
\centering
\begin{tabular}{|c c c c c|}
 \hline
 hipcentre & spine & head & ... &  shouldercentre\\ [0.5ex] 
 \hline
1.684013855 & 1.680728421 & 1.694426789 &...& 1.694013087 \\
1.687542467 & 1.680152844 & 1.689994086 &...& 1.690999615 \\
1.688034792 & 1.680716982 & 1.6897329 &...& 1.692213936 \\
1.688174173 & 1.680867813 & 1.690434507 &...& 1.69244093 \\
1.688240062 & 1.680919118 & 1.690826397 &...& 1.692456947 \\
1.688247039 & 1.681416153 & 1.690978133 &...& 1.693617492 \\
1.688279093 & 1.681536837 & 1.691291841 &...& 1.693789355 \\
1.688287023 & 1.681549649 & 1.691346302 &...& 1.69374882 \\
1.688258504 & 1.681548505 & 1.691332804 &...& 1.693702386 \\
1.688319709 & 1.681569607 & 1.691362489 &...& 1.693638131 \\
...&...&...&...&...\\ [1ex] 
 \hline
\end{tabular}
\caption{An excerpt of the first few lines of the 20-columned time series data on locations of the 20 joints of the body recorded at different time points, as Patient~3071 played the exergame \textit{Airplane}.}
\label{table:1}
\end{table}

\section{Learning the recovery trajectories}
\label{sec:implementation}
In order to accomplish the learning of the recovery trajectories of
patients whose exergame playing performance is recorded on the e-platform {\textit{MIRA}},
we will first learn the graph of the time series data (on
joint-locations) generated as a patient plays an exergame. Thereafter, we will
compute the inter-graph distance/divergence between the posterior
probabilities of the pair of graphs that are learnt given the time series data
on the 20 joint-locations, generated between two successive instances of
playing the exergame. We will estimate the correlation structure of such a time series
dataset, and then, conditional on that, we will compute the closed-form marginal
posterior of any edge of the graph variable, (see Equation~\ref{eqn:2}) to enable sampling from such a
marginal. 

\subsection{Computing correlation and partial correlation}
To define the inter-column correlation matrix
$\bSigma_{i,k,j}^{(C)}$, we compute the unbiased estimate of the correlation
between each pair of columns of the time series data on the 20
join-locations. Thus, this inter-column correlation matrix is $20\times
20$-dimensional. We will then transform each estimated correlation to the
partial correlation, using Eqn.~\ref{eq:1}, to compute the partial correlation
matrix $\bPsi_{i,k,j}^{(C)}$. Below, we present an excerpt of the inter-column
(i.e. the inter-joint-location) correlation matrix estimated from the time
series data generated by Patient~3071 playing \textit{Airplane}, i.e.

$$\bSigma_{3071,Airplane,1}^{(C)} = $$ 
\[
\begin{bmatrix}
1 & 0.330917541 & -0.26122886 &...& -0.165895188\\
0.330917541 & 1 & 0.68856342 &...& 0.870462199\\
-0.26122886 & 0.68856342 & 1 &...& 0.874589366\\
-0.319381464 & 0.544254133 & 0.692496065 &...& 0.730793308\\
-0.022265821 & 0.128763652 & 0.183350249 &...& 0.151056752\\
0.16721271 & -0.094951013 & 0.02250426 &...& -0.164855981\\
-0.051876216 & -0.138002972 & -0.006260628 &...& -0.100540177\\
\vdots & \vdots & \vdots & \ddots & \vdots\\
-0.165895188 & 0.870462199 & 0.874589366 &...& 1
\end{bmatrix}
\]

Then using this inter-joint-location correlation matrix, we compute the partial correlation matrix $\bPsi_{3071,Airplane,1}^{(C)}$, as:

$$\bPsi_{3071,Airplane,1}^{(C)}=$$
\[
\begin{bmatrix}
1 & 0.229345539 & -0.150743042 &...& -0.201536622\\
0.229345539 & 1 & -0.12561909 &...& 0.941688165\\
-0.150743042 & -0.12561909 & 1 &...& 0.299565913\\
0.041884517 & 0.13053044 & 0.210506791 &...& -0.023659251\\
0.024567646 & 0.052365098 & -0.128422074 &...& -0.018381679\\
0.067497723 & 0.002226935 & 0.259867599 &...& -0.070459859\\
-0.096152059 & -0.023513329 & -0.111474649 &...& 0.069071476\\
\vdots & \vdots & \vdots & \ddots & \vdots\\
-0.201536622 & 0.941688165 & 0.299565913 &...& 1
\end{bmatrix}
\]

\subsection{Rejection Sampling of an edge from the edge marginal}
\label{sec:rejection sampling}
We draw 50,000 samples of the edge variable $G_{s,s^{/}}$ that joins the
$s$-th and $s^{/}$ nodes of the random graph variable, where the $s$-th node
of this RGG
is associated with the location of the $s$-th joint in the body of the
patient, as they play an exergame; ($s=1,\ldots,20$).
The edge $G_{s,s^{/}}$ is sampled from its closed-form marginal
posterior $m(G_{s,s^{/}}\vert \psi_{s,s^{/}})$ that is articulated in
Equation~\ref{eqn:2}, and the sampling is performed using Rejection
Sampling. The binary edge variable $G_{s,s^{/}}$ attains a value
$g_{s,s^{/}}^{(r)}$ in the $r$-th sample.

In our implementation of Rejection Sampling, we use a Uniform proposal
density, (while we have experimented with a Bernoulli proposal as well), and we accept the proposed edge if and only if the acceptance ratio
exceeds or equals $u$ where $U=u$, with the Uniform random variable $U \sim U[0,1]$. The acceptance ratio of the $r$-th
iteration is
$$
\displaystyle{\frac{m(g^{(r)}_{s,s^{/}}\vert \psi_{s,s^{/}})}{C q(s,s^{/})}},
$$
with the constant $C$ chosen to
ensure that ${C q(s,s^{/})} \geq m(G_{s,s^{/}}\vert \psi_{s,s^{/}})$, such
that Rejection sampling is valid. 
Details of the aforesaid experimentation with the alternate Bernoulli proposal
density are discussed in Section~\ref{sec:robustness}.

For the example case of the patient with ID3071, playing the game
\textit{Airplane} 6 times, we use Rejection sampling to draw a sample of size
50,000 of the edge variable $G_{s,s^{/}}$, $\forall s^{/} > s; s = 1,2,\ldots,
19$, on each of the 6 instances. In other words, we learn realisations of
the graph variable using the time series (on joint locations) data generated
by this patient playing this exergame on these 6 occasions.

\subsection{Inter-graph distance and recovery trajectories: implementation details}
For the $i$-th patient playing the $k$-th exergame on the $j$-th instance, we compute the posterior
$\pi({\cal G}_{i,k,j}(\bV)\vert {\bf D}_{i,k,j})$ of the random graph variable
given the time series (on joint locations) data $\forall j=1,\ldots, N_{i,k}^{(max)}$. Then
we compute the inter-graph distance/divergence between this posterior of the graph variable learnt using the data
generated for the $j$-th instance, and that using the data obtained from this
patient playing this exergame on the
$(j-1)$-th instance, $\forall j=2,\ldots, N_{i,k}^{(max)}$. Such inter-graph distance/divergence is computed in
accordance with the methodology discussed in Section~\ref{sec:inter-graph
  distance}.

Table~\ref{table:5} shows the Hellinger distance and Kullbeck-Leibler
divergence between the posterior of the graph variable given the partial
correlation structure of the time series data generated when the patient with ID3071 plays
the exergame \textit{Airplane} on six successive instances. Table~\ref{table:6} shows
the corresponding MRSs that are computed using the distance/divergence, using
the methodology discussed in Section~\ref{sec:recovery trajectories MRS}
together with the points originally assigned by the e-platform {\textit{MIRA}}.

\begin{table}[H]
\centering
\begin{tabular}{|c c c|}
 \hline
 Successive instance pairs & Hellinger distance & Kullbeck-Leibler Divergence \\ [0.5ex] 
 \hline
(1,2) & 0.10604219  & 0.10995603 \\
(2,3) & 0.10589288  & 2.5307512e-5 \\
(3,4) & 0.0040824151  & 2.4243133e-6 \\
(4,5) & 0.017579886  & 2.3849898e-3 \\
(5,6) & 0.022596656  & 1.1390906e-3 \\[1ex] 
 \hline
\end{tabular}
\caption{Hellinger distance and Kullbeck-Leibler divergence between posteriors of the random graph variable learnt on
  successive instances, given time series (on joint locations) data generated by
  patient with ID3071 playing the exergame \textit{Airplane} on six instances.}
\label{table:5}
\end{table}

\begin{table}[H]
\centering
\begin{tabular}{|c c c c|}
 \hline
 Instance & MRS via Hellinger & MRS via KL & Points/scores assigned by e-platform {\textit{MIRA}}\\ [0.5ex] 
 \hline
1 & 0.05302109 & 0.05497801 & 230 \\
2 & 0.15906328 & 0.16493404 & 255 \\
3 & 0.26495617 & 0.16495935 & 290 \\
4 & 0.26903858 &  0.16496177 & 360 \\
5 & 0.28661847 & 0.16734676 & 395 \\
6 & 0.30921512 & 0.16848585 & 315 \\[1ex] 
 \hline
\end{tabular}
\caption{MRS calculated using the Hellinger distance and KL
Divergence displayed in Table~\ref{table:5}, and points assigned by e-platform
{\textit{MIRA}}, for patient with ID3071 playing the exergame {\it{Airplane}} on six instances}
\label{table:6}
\end{table}

\begin{figure}[H]
    \centering
    \includegraphics[width = \textwidth]{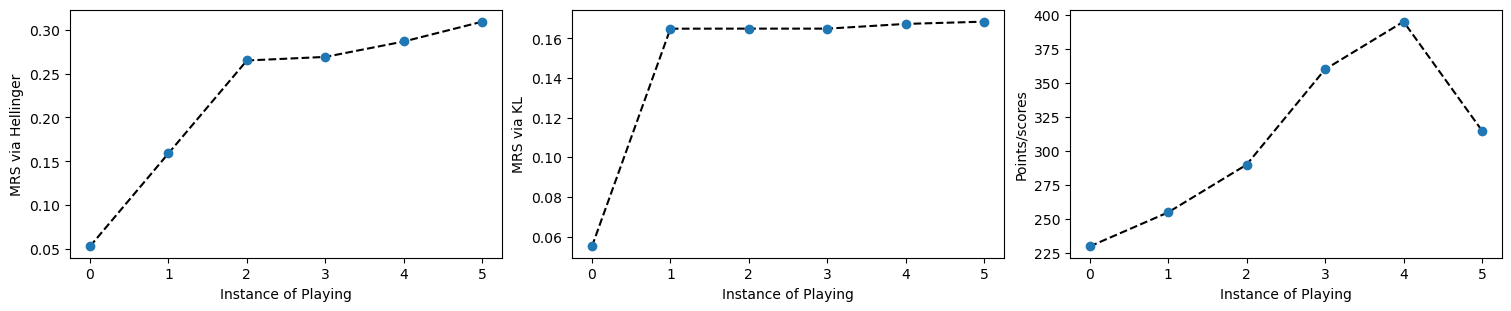}
    \caption{Figure displaying plots of performance of patient ID3071 playing
      the exergame {\it{Airplane}}, against the index of the instance of
      playing this exergame. On the left, the plot of MRS computed using
      Hellinger distance is plotted against the index of the instance of
      playing. In the middle panel, the MRS computed using Kullback–Leibler
      divergence is plotted, while on the right, points assigned by the {\textit{MIRA}} e-platform for playing the game on 6 instances, are plotted against the instance index.}
    \label{fig:3071_airplane_3trajectories}
\end{figure}

We visualise the recovery trajectories as the variation of the computed MRS
against the index of the instance, when a given patient plays a given
exergame. Here, the MRS is computed using either the Hellinger distance or the KL
divergence, as per Section~\ref{sec:recovery trajectories MRS}. We also
compare our learnt MRS - and thereby the learnt recovery trajectory - against
the points that are assigned by the e-platform, across instances in which a
game is played. Figure~\ref{fig:3071_airplane_3trajectories} shows examples of
these recovery trajectories for patient with ID3071, playing the exergame {\it{Airplane}} on six
occasions.

\section{Results}
\label{sec:results}
In this section, we report the results obtained from computing values of the
parameter that embodies all relevant information about the differences between
two disparately-long multivariate time-series datasets on 20 monitored joints
of a patient$^{,}$s skeletal framework, produced as the patient plays an
exergame, on successive instances. Our parametrisation of such differences
between a pair of such successively produced time series datasets, informs on
the progress of the patient in recovering their movement ability between such
successive playing of the exergame. Above, we have discussed the
conceptualisation of this parametrisation as the distance/divergence between
the posterior probabilities of random graphs that are learnt given the
respective time series data. We conduct such inter-graph distance computation
relevant to each patient in the cohort, playing each of the games on
successive instances. As discussed in Section~\ref{sec:implementation}, we
will compute this inter-graph distance for each pair of graphs - learnt given
the successive pairs of time series data - to ultimately construct a recovery
trajectory for each patient playing an exergame.

\subsection{Realisations of the graph variable at chosen $\tau$}
\begin{figure}[H]
     \centering
     \begin{subfigure}
         \centering
         \includegraphics[width=0.3\textwidth]{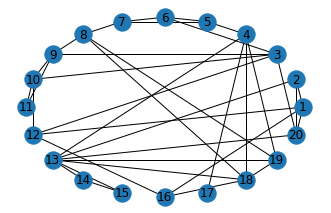}
     \end{subfigure}
     \hfill
     \begin{subfigure}
         \centering
         \includegraphics[width=0.3\textwidth]{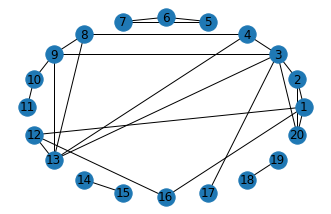}
     \end{subfigure}
     \hfill
     \begin{subfigure}
         \centering
         \includegraphics[width=0.3\textwidth]{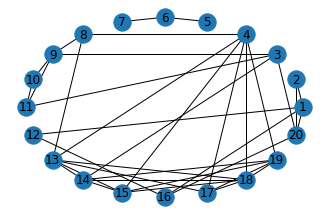}
     \end{subfigure}
    \hfill
     \begin{subfigure}
         \centering
         \includegraphics[width=0.3\textwidth]{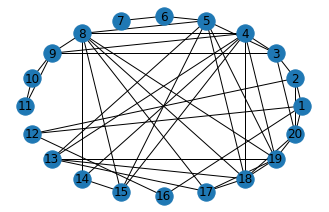}
     \end{subfigure}
     \hfill
     \begin{subfigure}
         \centering
         \includegraphics[width=0.3\textwidth]{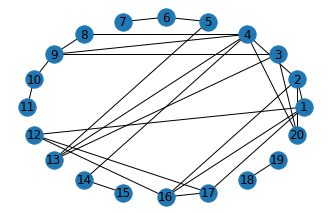}
     \end{subfigure}
     \hfill
     \begin{subfigure}
         \centering
         \includegraphics[width=0.3\textwidth]{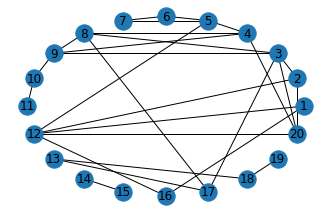}         
     \end{subfigure}   
     
    \caption{Realisations of graphs learnt at a chosen $\tau$ of 0.2, given data recorded for patient ID3071 playing exergame {\it{Airplane}} on
six successive instances, in the order of the instance of playing. 
}
\label{fig:3071_graph_airplane}
\end{figure}

\begin{figure}[H]
     \centering
     \begin{subfigure}
         \centering
         \includegraphics[width=0.3\textwidth]{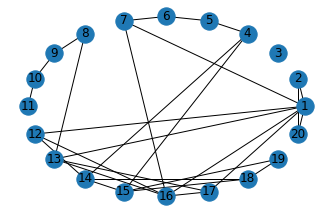}
     \end{subfigure}
     \hfill
     \begin{subfigure}
         \centering
         \includegraphics[width=0.3\textwidth]{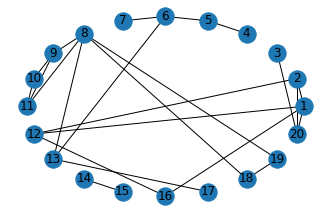}
     \end{subfigure}
     \hfill
     \begin{subfigure}
         \centering
         \includegraphics[width=0.3\textwidth]{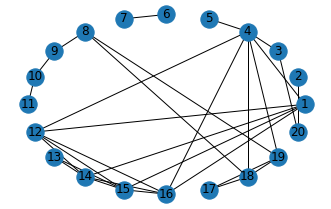}
     \end{subfigure}
     \hfill
     \begin{subfigure}
         \centering
         \includegraphics[width=0.3\textwidth]{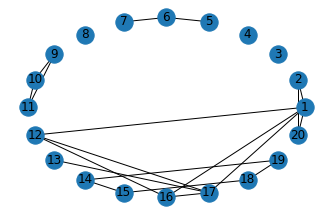}
     \end{subfigure}
       \hfill
     \begin{subfigure}
         \centering
         \includegraphics[width=0.3\textwidth]{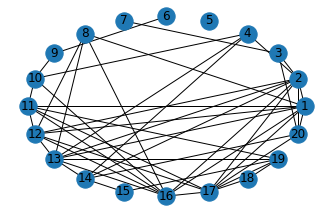}
     \end{subfigure}
        \hfill
     \begin{subfigure}
         \centering
         \includegraphics[width=0.3\textwidth]{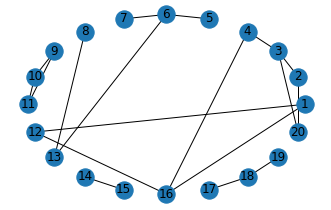}
     \end{subfigure}

    \caption{Realisations of the graph variable, learnt at $\tau=0.3$, given data recorded for the patient ID3311 playing exergame {\it{Follow}}, on
35 successive instances - of which, results of playing on 6 instances are presented here, in the order of the instance of playing. 
}
\label{fig:3311_graph_follow}
\end{figure}

In Figure~\ref{fig:3071_graph_airplane} we present the realisations of the
graph variable, learnt at a $\tau$ of 0.2, given the time series (on joint locations) datasets
that are produced, as patient ID3071 plays \textit{Airplane} on six different
instances. Reaisations learnt at $\tau=0.3$, given data produced by the patient ID3311 playing exergame
\textit{Follow}, are presented in Figure~\ref{fig:3311_graph_follow}.

\subsection{Recovery Trajectories and MRS}
\begin{figure}[H]
     \centering
     \begin{subfigure}
         \centering
         \includegraphics[width=\textwidth]{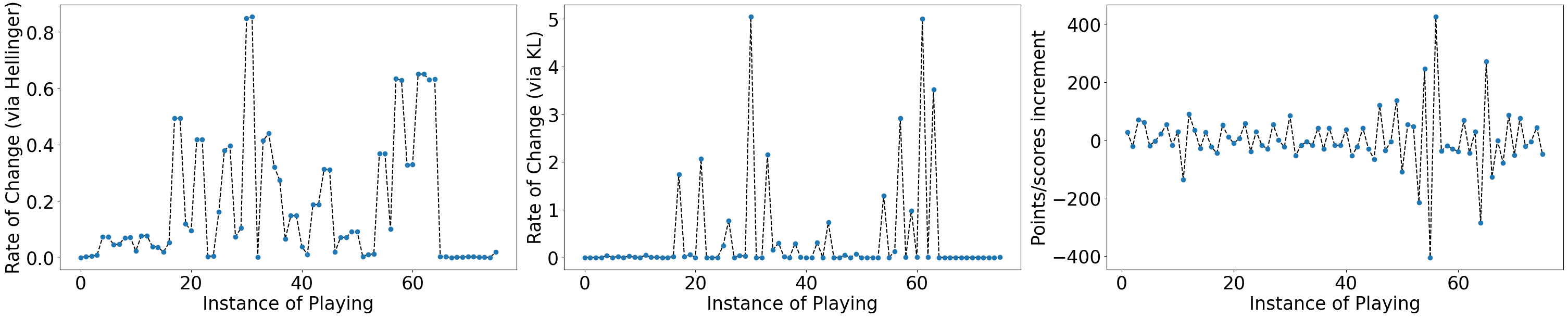}
     \end{subfigure}
     \begin{subfigure}
         \centering
         \includegraphics[width=\textwidth]{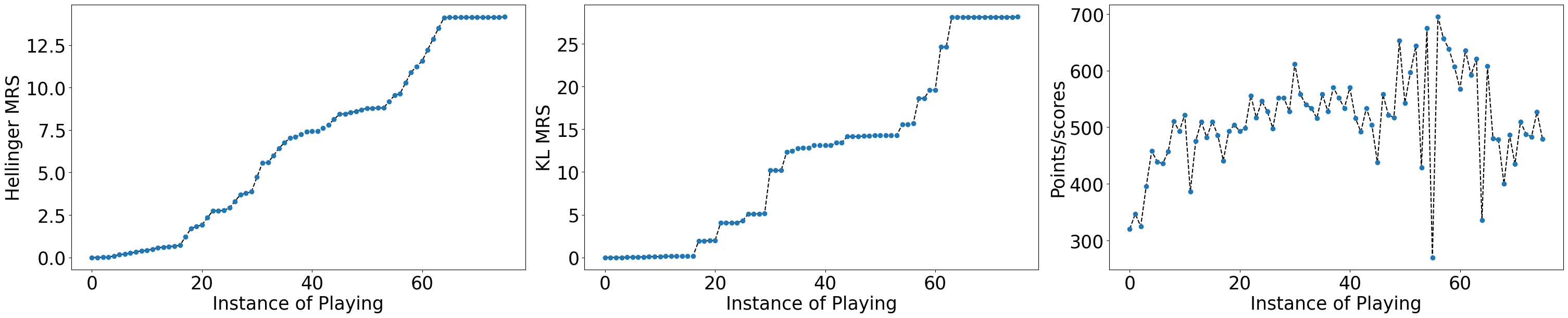}
\end{subfigure}
    \caption{Results for the patient with ID3085, playing the exergame {\it{Catch}}, 76
      times. The panels in the first row display plots of rate of change of
      MRS (left and middle) and that of the points assigned by the e-platform
      {\textit{MIRA}} (right). at various instances. The panels in the bottom
      row display the MRS (left and middle) and the points assigned by
      {\textit{MIRA}}. In the first column, Hellinger distance
      has been employed to compute the distance between posteriors of the
      graph variables learnt using data generated at successive instances of
      playing the exergame. The second column is similar to the first, except
      here, results are shown for the inter-graph divergence computed using
      the Kullback-Leibler divergence.}
\label{fig:3085_Catch}
\end{figure}

\begin{figure}[H]
     \centering
     \begin{subfigure}
         \centering
         \includegraphics[width=\textwidth]{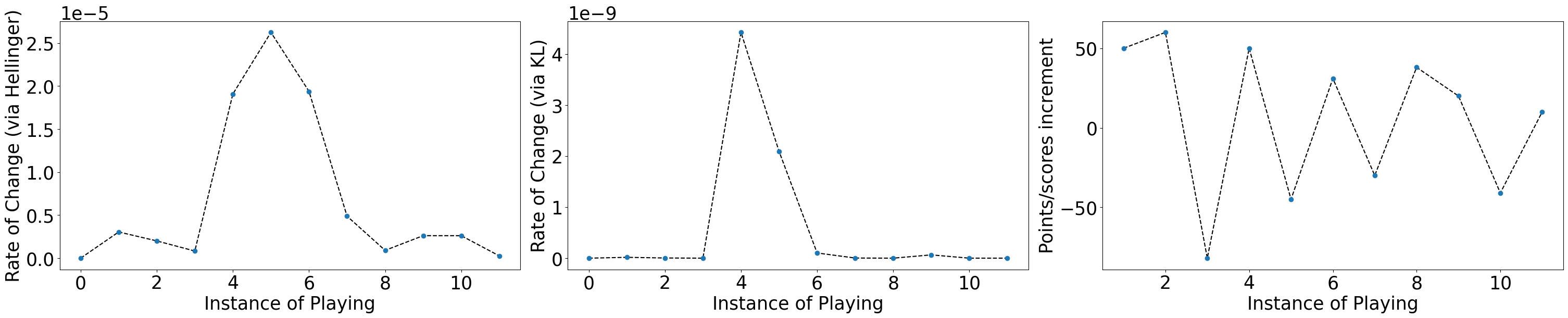}
     \end{subfigure}
     \begin{subfigure}
         \centering
         \includegraphics[width=\textwidth]{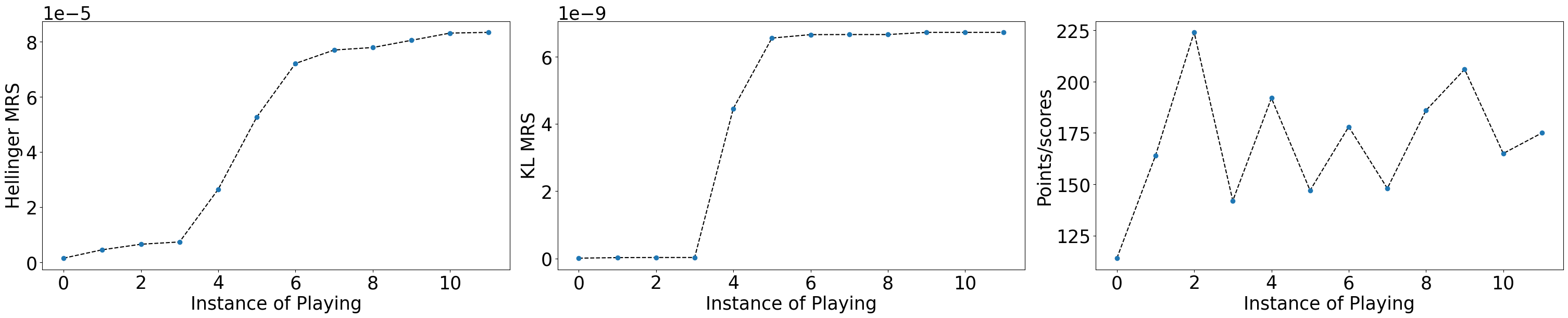}
\end{subfigure}
    \caption{Same as for Figure~\ref{fig:3085_Catch}, with results displayed for the patient ID3327 playing the exergame {\it{IzzytheBee}} 12 times. }
\label{fig:3327_IzzytheBee}
\end{figure}

Having learnt the posteriors of the graph variable given the time series (on joint
locations) data generated by the $i$-th patient playing the $k$-th exergame, on
all relevant instances, we proceed to construct the recovery trajectories for
this patient, as they play a given exergame. The following figures show the
evolution in the learnt MRS as the patient repeatedly plays an exergame,
across all instances of playing it. In Figure~\ref{fig:3085_Catch} and
Figure~\ref{fig:3327_IzzytheBee} we display recovery trajectories of the
patient ID3085 playing the exergame \textit{Catch} 76 times, and the patient
ID3327 playing exergame \textit{IzzytheBee} 12 times. For each patient, we show the comparison of the rate of change of MRS, and the MRS - computation of which is discussed in
Section~\ref{sec:recovery trajectories MRS} - using the Hellinger distance function; KL
divergence; as well as the scores (points gained at the end of playing an exergame) provided by {\textit{MIRA}}.

From Figure~\ref{fig:3085_Catch} and Figure~\ref{fig:3327_IzzytheBee}, it can
be seen that MRSs computed using Hellinger/Kullbeck-Leibler, represent clearly
visualised recovery trajectories, based on robust learning of the performance
indicator of recovery. Indeed, this learning is robust to variation in the
number of instances of playing an exergame, as well as in the time taken for a
patient to play an exergame.
\begin{remark}
Evolution of the points provided by the e-platform {\textit{MIRA}} does not
offer a meaningful trend. Instead, the inbuilt point assignment facility on
{\textit{MIRA}} offers wide variations in the performance indicator. Thus, the feedback on recovery offered on the basis of the points assigned
by {\textit{MIRA}}, is far less clear than that provided by MRS values.
\end{remark}
It is worth noting that the level of difficulty of the exergames changes over time, giving rise to a temporary depression in performance, when the patient moves to a harder level. 

In general, recovery trajectories computed using the Hellinger distance and KL
divergence measures, roughly agree with each other. However by the definition
of distance, the Hellinger distance will yield the absolute of the difference
between the scores attained by a patient when playing a game on two
successive instances; therefore, the MRS computed using the Hellinger distance
measure is never smaller on one instance, compared to the MRS computed at the
previous instance. On the other hand, the Kullbeck-Leibler divergence allows a
negative value for the divergence between the posteriors of graph models
learnt using (partial correlation of) the joint-location time series data
generated on successive instances of playing a game - which is the rate of
change in the MRS between the two successive instances. Therefore, the MRS
computed using the KL divergence function on one instance can be smaller than
that in the previous. Then it follows that the recovery trajectory learnt
using the KL divergence measure will not adhere to local monotonicity, (as
noted in the recovery trajectories of the subject ID3147 playing game
\textit{Follow}, the subject ID3327 playing game \textit{Nimble} and so on).
\begin{remark}
We find it an informative result that most of the recovery trajectories learnt with
both the distance and divergence measures concur, indicating that most of the
patients in the cohort, avail of improving movement ability, with increasing
instances of playing the game.
\end{remark}

\begin{remark}
In applications to learning recovery trajectories using generic datasets, we
recommend the usage of KL-divergence over Hellinger distance, based only on
the criterion that the KL-divergence
can produce both positive and negative rates of change of the MRS, while such a
rate is essentially positive when the MRS is computed using the Hellinger
distance function.
\end{remark}

\subsection{Robustness of the graph learning}
\label{sec:robustness}

\begin{figure}[H]
     \centering
     \begin{subfigure}
         \centering
         \includegraphics[width=0.4\textwidth]{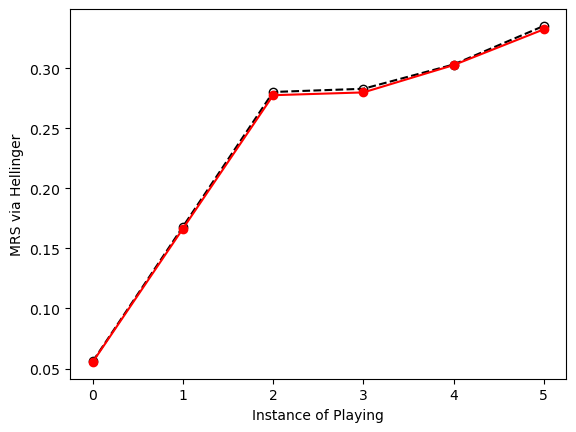}
     \end{subfigure}
     \begin{subfigure}
         \centering
         \includegraphics[width=0.4\textwidth]{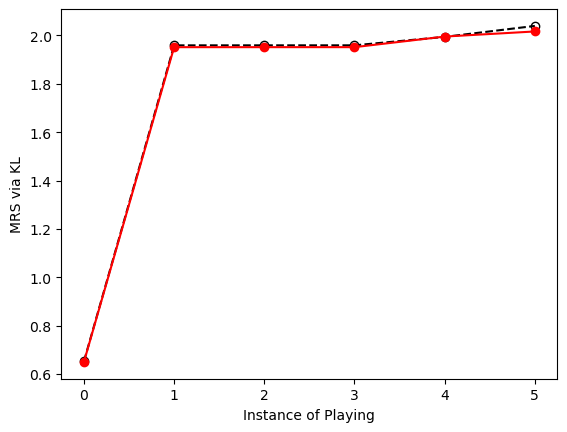}
\end{subfigure}
    \caption{Comparison of recovery trajectories of patient ID3071
      playing the exergame {\it{Airplane}} on 6 instances, learnt using Uniform proposal (in
      black dashed lines and unfilled circles) and Bernoulli proposal
      (in red - or grey in the monochromatic version of the articles - solid
      lines and red circles). Results using the Hellinger distance function
      are displayed in the left panel while results using KL-divergence are
      shown on the right. 
    }
\label{fig:3071_Airplane_robust}
\end{figure}

We checked for the robustness of graph learning to variation in the
proposal density used in the undertaken Rejection Sampling, by learning the
recovery trajectories of the patient ID3071, playing the game \textit{Airplane} on six instances, using both a Bernoulli and a Uniform proposal in our
implementation of Rejection sampling. For this example
case, we drew a sample of size $500,000$ for $g_{s,s^{/}}$, $\forall s^{/}> s;
s=1,2,\ldots,19$, once with a $Bernoulli(|\psi_{s,s^{/}}|)$ proposal, and separately,
with a $Uniform[0,1]$ proposal. We computed the inter-graph distance thereafter, to
ultimately learn the recovery trajectories. The learnt trajectories are
plotted in Figure~\ref{fig:3071_Airplane_robust} for inter-graph distance
learnt using the Helllinger distance measure and the KL divergence
measure. Near-concurrence of the learnt trajectories testify to the robustness
of our learning to changes in details of the undertaken Rejection Sampling.

\subsection{More Recovery Trajectories}
\begin{figure}[H]
     \centering
     
          \begin{subfigure}
         \centering
         \includegraphics[width=0.45\textwidth]{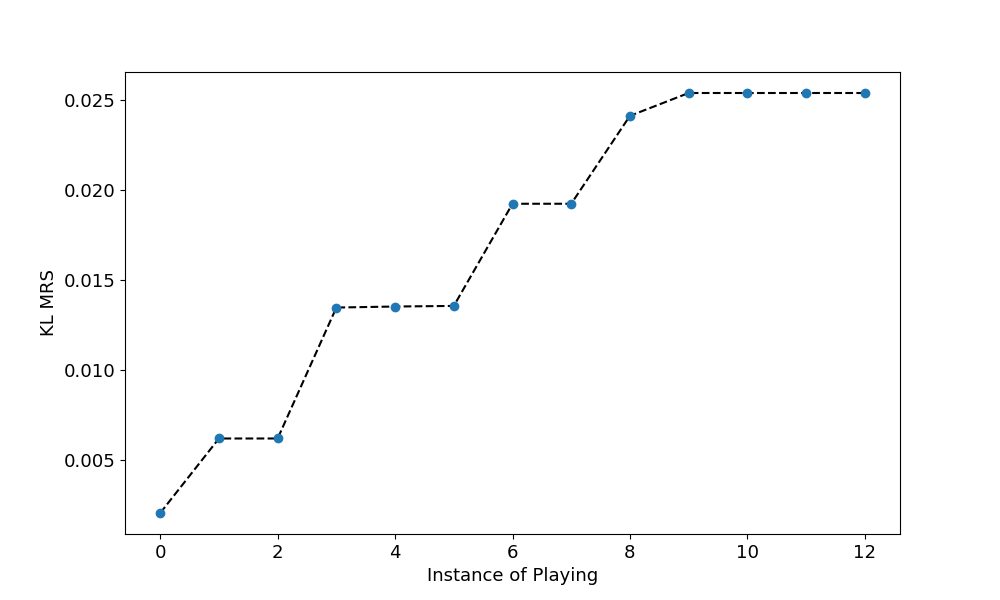}
     \end{subfigure}
       \hfill
       \begin{subfigure}
         \centering
         \includegraphics[width=0.45\textwidth]{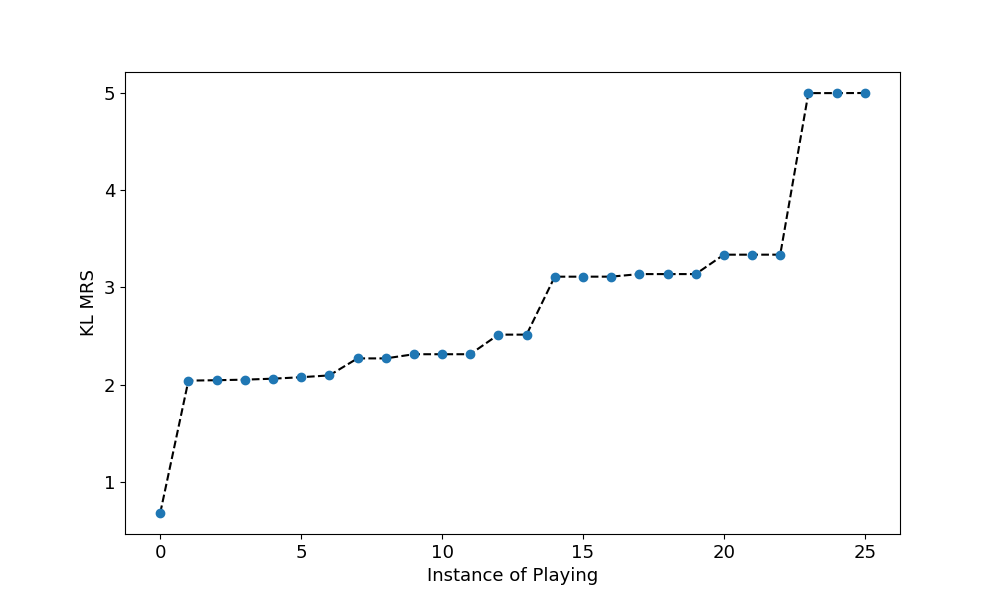}
     \end{subfigure}
     \hfill
         \begin{subfigure}
         \centering
         \includegraphics[width=0.45\textwidth]{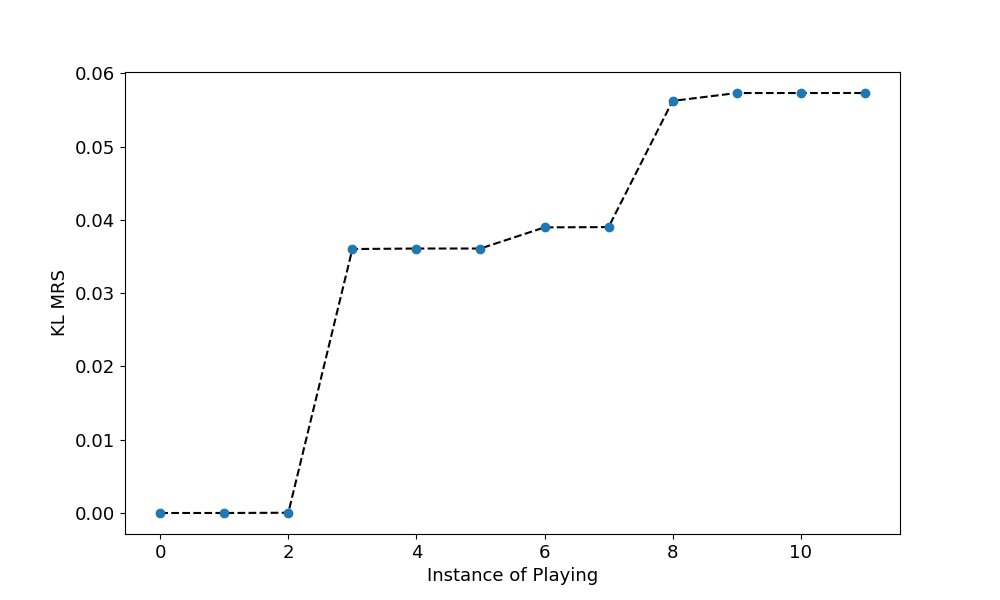}
     \end{subfigure}
       \hfill 
     \begin{subfigure}
         \centering
         \includegraphics[width=0.45\textwidth]{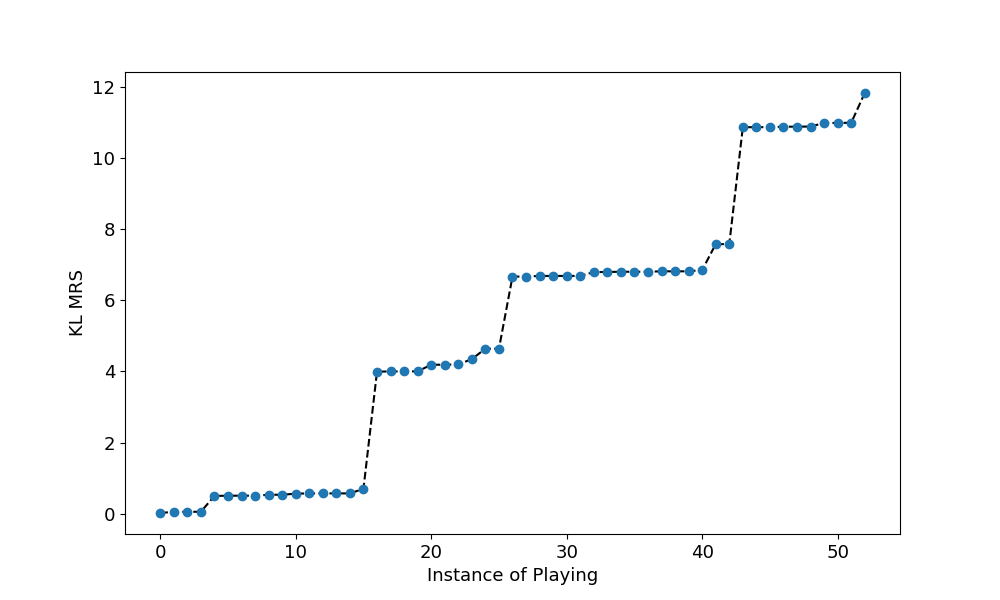}
     \end{subfigure}
     \hfill
     \begin{subfigure}
         \centering
         \includegraphics[width=0.45\textwidth]{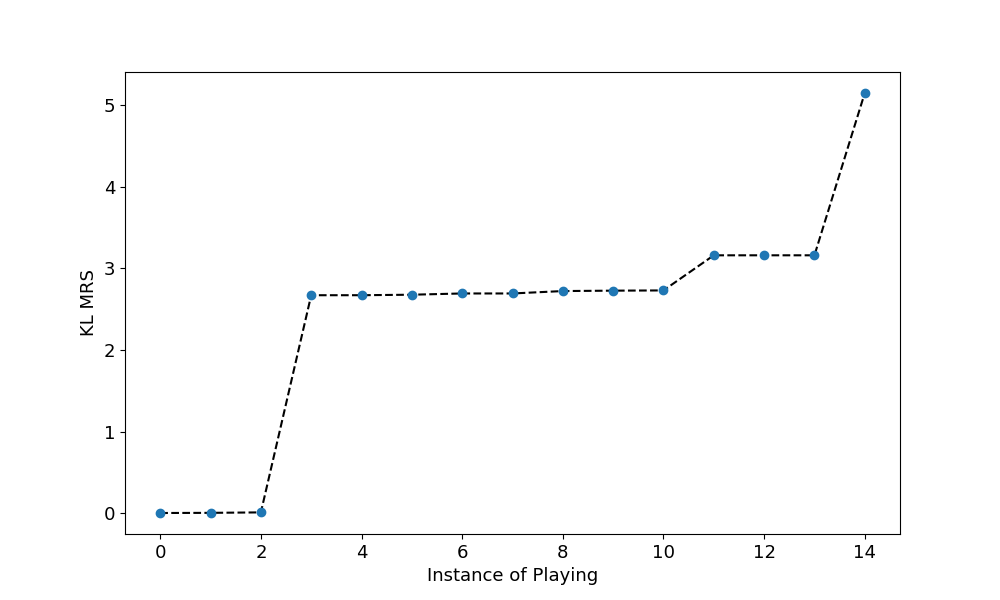}
     \end{subfigure}
       \hfill
     \begin{subfigure}
         \centering
         \includegraphics[width=0.45\textwidth]{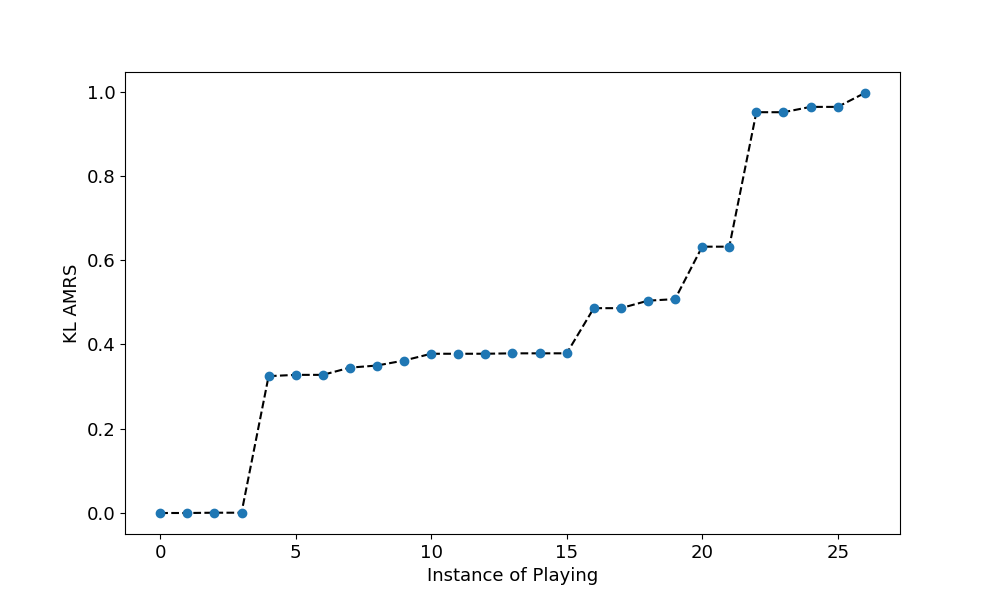}
     \end{subfigure}
        \hfill
     \begin{subfigure}
         \centering
         \includegraphics[width=0.45\textwidth]{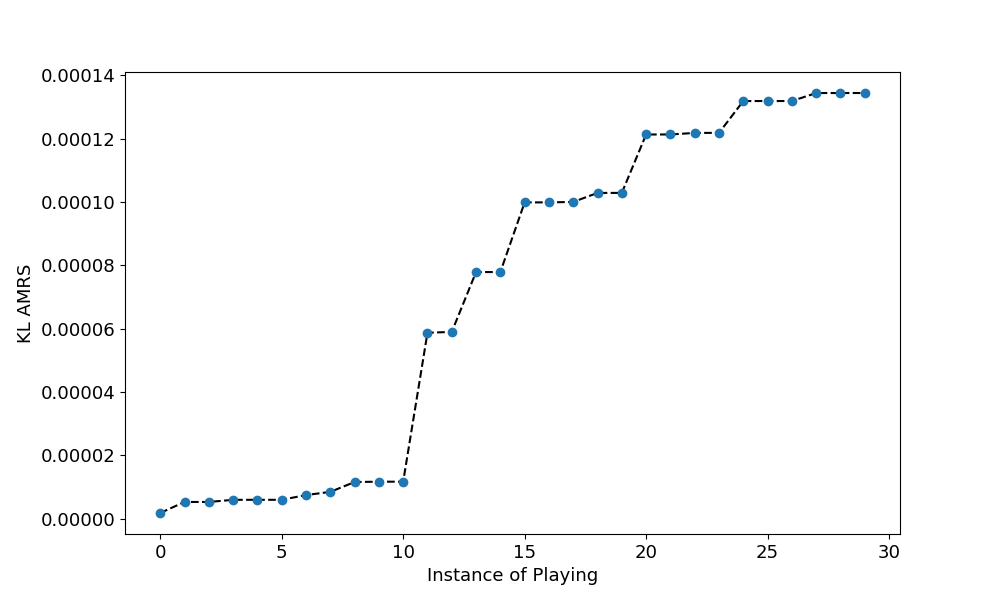}
    \end{subfigure}
            \hfill  
     \begin{subfigure}
         \centering
         \includegraphics[width=0.45\textwidth]{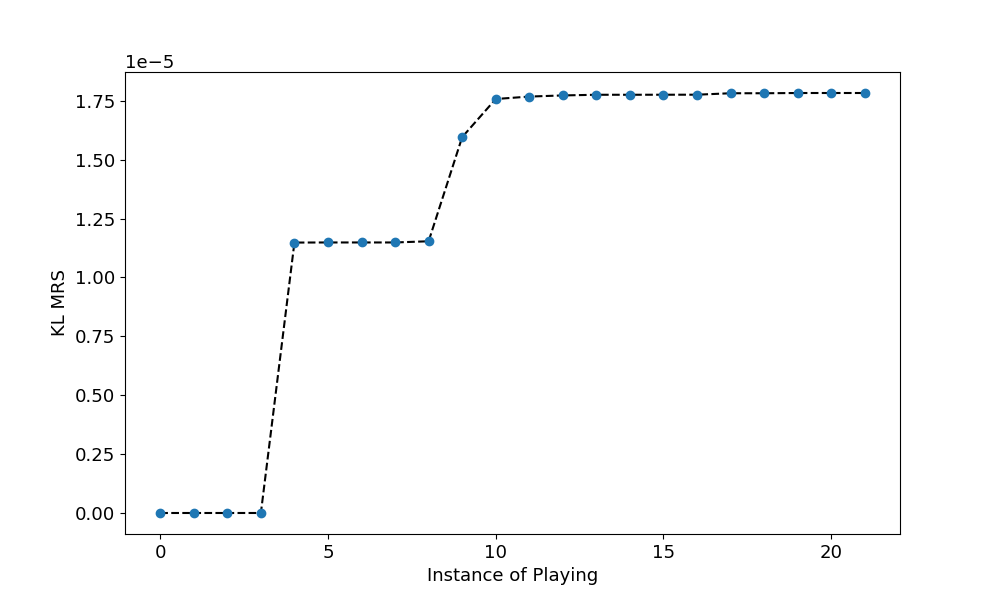}
    \end{subfigure}

    \caption{Recovery trajectories of patients, computed using KL.}
\label{fig:more_kl}
\end{figure}

Figure~\ref{fig:more_kl} presents a collection of more recovery trajectories
learnt for different patients playing different exergames, multiple times,
using MRS computed using
the Kullbeck-Leibler divergence measure.

In general, from our calculated recovery trajectories, most of the patients
show an improvement in the physical rehabilitation during the training (which
is not that evident from the statistics provided by {\textit{MIRA}}
platform). This finding is also supported by the MRI profile scans and patient
functional assessment taken at the beginning and the end of the training
programme
that showed significant improvement in motor function \cite{Firwana}. Our method provides an easy way to visualise and interpret the recovery information of such patients.

\subsection{Discussion of results}
\label{sec:Discussion}

\begin{figure}[H]
    \centering
    \includegraphics[width=\textwidth]{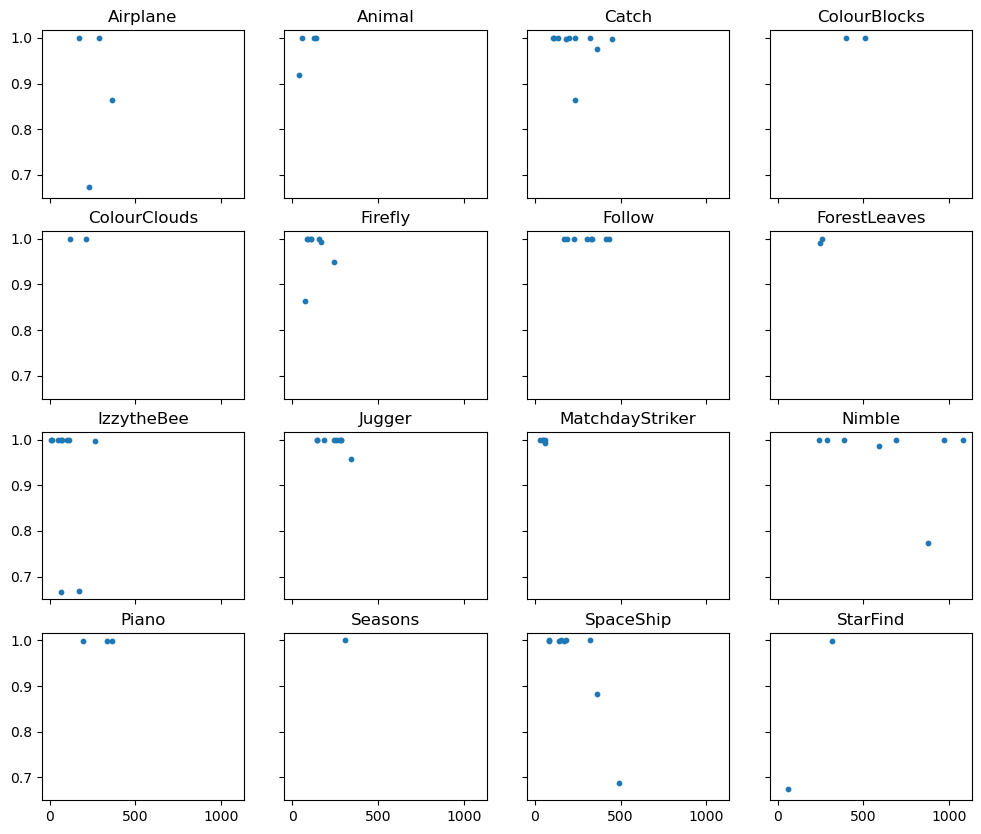}
    \caption{Plots of computed recovery parameter of all patients playing a
      particular exergame, (noted in the panel header), against the points
      scored by a patient on the MIRA e-platform, in the first instance that they play the considered exergame. The recovery parameters are plotted only for those patients who played this exergame in at least four instances}
    \label{fig:4by4_games}
\end{figure}

We now collate the results of learning recovery trajectories for this cohort
of patients, to advance recommendations as to the “best” suited exergame for a
patient with a given level of mobility-impediment at the stage when the
physical rehabilitation programme is just about to begin. This calls for the
identification of the pre-rehab mobility level, and of a (scalar-valued)
parameter that measures recovery. We discuss these below.
\begin{enumerate}
\item[---]The best proxy for the pre-rehab mobility level of the $i$-th patient is
  the automated score ${\cal S}_{i,k}(1)$ that they obtain on the e-platform {\textit{MIRA}},
  when undertaking the $k$-th exergame in the first instance of playing this
  exergame. The lower is ${\cal S}_{i,k}(1)$, the higher is the mobility impediment faced by the $i$-th patient. 
\item[---]The recovery trajectory itself is the most appropriate recovery
  measure, but it is a vector-valued variable (that approximates a
  function). Here, we implement the recovery trajectory that results from
  the MRS computed using the KL-divergence measure. We use such a recovery trajectory to compute the scalar-valued recovery parameter
  $\alpha_{i,k}$ of the $i$-th patient playing the $k$-th exergame, as the
  (normalised) range of such MRS values attained by this patient over the full
  course of
  them playing this exergame: $$\alpha_{i,k} \coloneqq \frac{{\cal
      MRS}_{i,k}^{(KL)}(N_{i,k}^{(max)}) - {\cal MRS}_{i,k}^{(KL)}(2)}{{\cal
      MRS}_{i,k}^{(KL)}(N_{i,k}^{(max)})}.$$ 
\end{enumerate}

We then plot $\alpha_{i,k}$ against ${\cal S}_{i,k}(1)$, for all those patients who
played the $k$-th game at least four times. The plots for the 16 relevant
exergames, \textit{Airplane, Animal, Catch, ColourBlocks, ColourClouds,
  Firefly, Follow, ForestLeaves, IzzytheBee, Jugger, MatchdayStriker, Nimble,
  Piano, Seasons, SpaceShip, StarFind} are shown for 10 patients in
Figure~\ref{fig:4by4_games}. There are two points that need to be appreciated:
\begin{enumerate}
\item between two exergames, the one that demonstrates a higher recovery parameter at lower initial scores, (i.e. for more severe mobility impediment), is the more efficient of the two;
\item we can use this figure to suggest the “best” exergames for patients with a given level of severity of mobility difficulties, as parametrised by the initial score. 
\end{enumerate}
From point~1 above, in the cohort we have analysed, the exergames with the highest
rehabilitating proficiencies are: {\it{IzzytheBee; MatchdayStriker; Animal}} - in
that order. Again, in this data, the exergames {\it{Catch; Follow; Jugger}}
appear to be the ones that are nearly equally proficient in rehabilitating
patients across a wide range of severity of mobility impediments. However,
{\it{Nimble}} appears to be a game with little discretionary capacity - all
patients, irrespective of their initial level of mobility deficiency, appear
to attain high recovery upon playing this exergame. So {\it{Nimble}} is not
recommended as a suitable exergame.

So a patient who suffered from severe mobility impediment at the start of their rehab programme - on the basis of the information that we have at hand - is best recommended to play {\it{IzzytheBee; MatchdayStriker; Animal}}, and perhaps {\it{Firefly}} and {\it{Spaceship}} as well. For those with less severe impediments,  {\it{Catch; Follow; Jugger}} are more suitable. Those with the least severe difficulties are recommended to play {\it{ColourBlocks; Catch; Follow}}.

It is worth mentioning that our approach could be both applied to any different similar platform or could be expanded beyond a virtual reality platform, to a simple set of exercises that could be done in a home environment.

\section{Conclusions}
\label{sec:conclusion}
Our method of learning recovery trajectories allows for patient-specific
recommendations of the optimal route for rehab, where said recommendations are
provided at an early stage, (i.e. before the physical rehab has started). In
a future contribution, we will be undertaking the prediction of the full
recovery trajectory of an individual patient. To perform such a prediction, the pre-requisite is a training dataset that would permit the (supervised) learning of the
relation between pre-rehab severity of mobility impediment in an individual
patient, and the recovery trajectory that marks this patient$^{,}$s recovery as
they play the $k$-th exergame, where $k$ indexes any of the exergames marked in
Figure~\ref{fig:4by4_games}. Such a supervised learning will enable the prediction of the recovery trajectory for a patient with a given level of mobility impairment, at the start of the therapeutic programme. Such a prediction will offer physicians and
therapists a robust patient-specific and early assessment of a given patient’s
course of recovery. Already with the results reported here, we offer robust
recommendations about the optimal exergames suitable for a patient with
categorised severity, where severity at the pre-rehab stage above, is proxied
by the patient’s initial score that is recorded (in an automated way) on the
e-platform {\textit{MIRA}} as the patient plays the exergame in question. In general applications, the initial state of mobility impediment could be informed upon otherwise/additionally as well.

We note that our learning of the recovery trajectories is founded on rigorous
mathematics, namely on the statistical distance/divergence between random
graph variables that are learnt given respective data slices. Each such dataset
comprises a time series of locations of 20 joints of the patient’s skeletal
frame, recorded while the patient undertakes an exergame on various
instances. Then the distance between the graphs learnt given data generated on
playing at the current instance and the previous, allows for the computation
of the MRS relevant to the current instance.

We forward this method as one that is relevant to the wider and topical quest
for early and patient-specific diagnosis. Additionally, this automated
protocol is easy to implement and is suitable for small cohort sizes;
prediction of the optimal exergame is fast; and crucially, the learning of
recovery trajectories by this method is robust and explainable. Given recent
developments in the identification of the ``optimal'' $\tau$ in a given
dataset, in a future contribution, we will compute the distance/divergence
between two RGG variables that are learnt given the correlation structure
of the respective datasets, using the identified optimal $\tau$
values in the relevant dataset.

\bibliographystyle{unsrt}

\end{document}